\documentclass[3p, review]{elsarticle}

\usepackage{lineno,hyperref}
\usepackage{booktabs}
\modulolinenumbers[5]

\usepackage[figuresright]{rotating}
\usepackage{amsopn}
\usepackage{amssymb}
\usepackage{amsmath}
\usepackage{subfig}
\usepackage{array}
\usepackage{tabularx}
\usepackage{multirow}
\usepackage{booktabs}
\usepackage{graphicx}
\usepackage{mathtools}
\usepackage{algorithmic}
\usepackage{algorithm}
\usepackage{listings}
\usepackage{setspace}
\usepackage{url}
\usepackage{float}
\usepackage{epsf} 
\usepackage{tikz}
\usepackage{amsthm}
\usepackage{color}
\usepackage{lineno}

\usepackage{multirow}
\usepackage{graphicx}
\usepackage[normalem]{ulem}
\useunder{\uline}{\ul}{}

\theoremstyle{definition}
\usepackage{url}

\usepackage{etoolbox}

\journal{Pattern Recognition}

\makeatletter
\patchcmd{\ps@pprintTitle}
 {Preprint submitted to}
 {Accepted by}
 {}{}
\makeatother
\begin{document}

\begin{frontmatter}

\title{CAGNet: Content-Aware Guidance for Salient Object Detection}


\author[a]{Sina Mohammadi\fnref{equal}\corref{Corresponding}}
\ead{sina.mhm93@gmail.com}
\fntext[equal]{Both the authors contributed equally.}

\author[a]{Mehrdad Noori\fnref{equal}\corref{Corresponding}}
\ead{me.noori.1994@gmail.com}
\cortext[Corresponding]{Corresponding authors.}

\author[a]{Ali Bahri}
\author[a]{Sina Ghofrani Majelan}
\author[b]{Mohammad Havaei}
    
\address[a]{School of Electrical Engineering\unskip, 
    Iran University of science and Technology\unskip, Tehran, Iran}
  	
\address[b]{
    Imagia Inc.\unskip, Montreal, Canada}

\begin{abstract}

Beneficial from Fully Convolutional Neural Networks (FCNs), saliency detection methods have achieved promising results. However, it is still challenging to learn effective features for detecting salient objects in complicated scenarios, in which i) non-salient regions may have "salient-like" appearance; ii) the salient objects may have different-looking regions. To handle these complex scenarios, we propose a Feature Guide Network which exploits the nature of low-level and high-level features to i) make foreground and background regions more distinct and suppress the non-salient regions which have "salient-like" appearance; ii) assign foreground label to different-looking salient regions. Furthermore, we utilize a Multi-scale Feature Extraction Module (MFEM) for each level of abstraction to obtain multi-scale contextual information. Finally, we design a loss function which outperforms the widely used Cross-entropy loss. By adopting four different pre-trained models as the backbone, we prove that our method is very general with respect to the choice of the backbone model. Experiments on six challenging datasets demonstrate that our method achieves the state-of-the-art performance in terms of different evaluation metrics. Additionally, our approach contains fewer parameters than the existing ones, does not need any post-processing, and runs fast at a real-time speed of 28 FPS when processing a $480\times480$ image.
\end{abstract}


\end{frontmatter}

\doublespacing

\section{Introduction}

Salient object detection aims at localizing the most interesting and prominent parts of an image. Moreover, it is an effective pre-processing step for numerous computer vision tasks such as image classification~\cite{flores2019saliency}, image segmentation~\cite{li2016robust, zhi2018saliency, cai2019saliency}, video segmentation~\cite{wang2015saliency}, image editing~\cite{chen2015improved, zhang2018novel} and object tracking~\cite{hong2015online}.

\begin{figure}
\begin{center}
\includegraphics[width=0.7\linewidth]{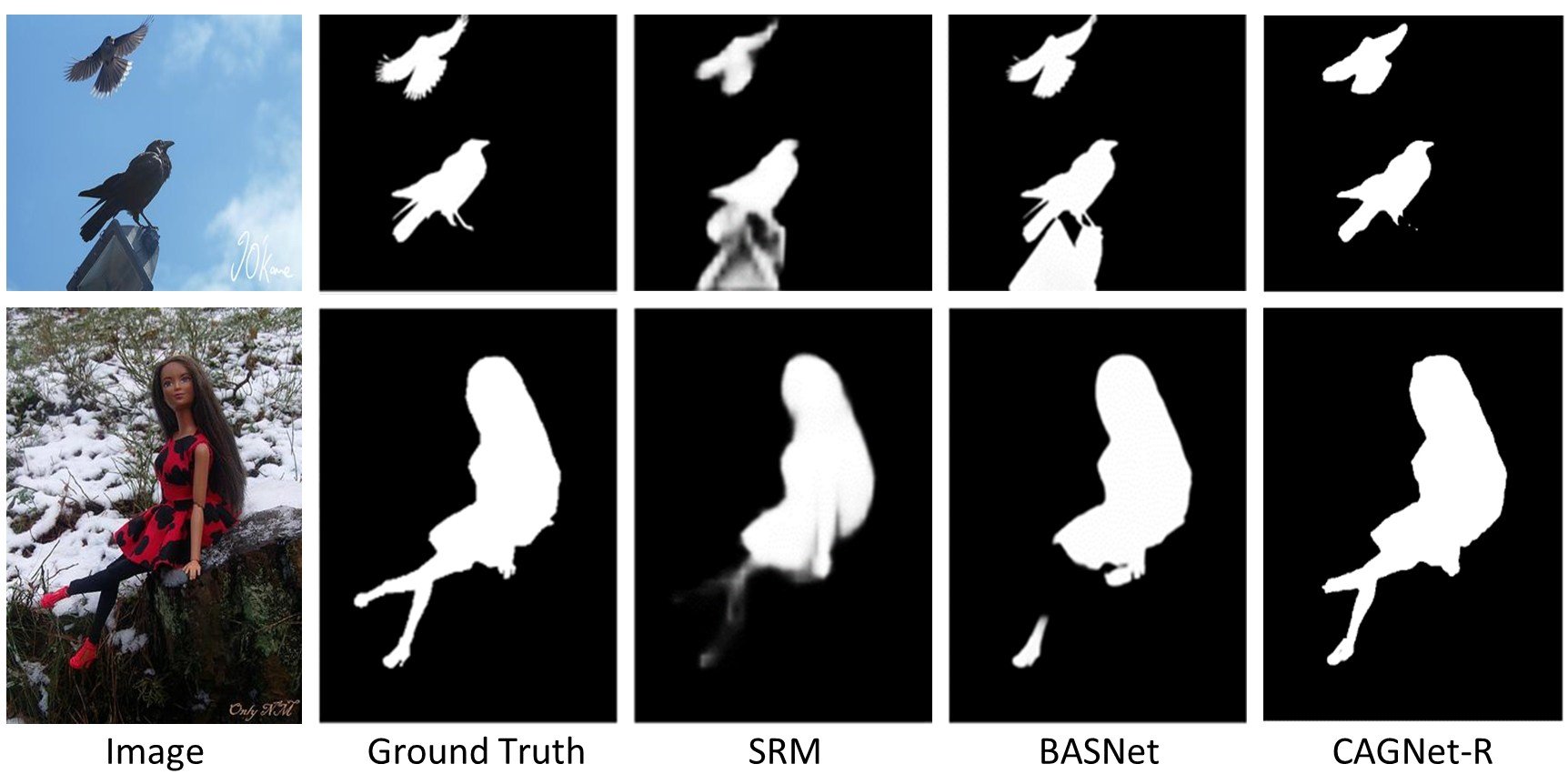}
\end{center}
\caption{Examples of complicated scenarios in salient object detection. In the first row, the triangular object has "salient-like" appearance. In the second row, the appearance of the feet of the doll is different from the rest of the doll. While both scenarios have caused confusion for two recent methods (BASNet~\cite{qin2019basnet} and SRM~\cite{wang2017stagewise}), our method (denoted as CAGNet-R) is capable of handling these complicated scenarios and generating more accurate saliency prediction.}
\label{fig:SM_first_page}
\end{figure}

Traditional approaches are mostly based on low-level cues and hand-crafted features. For example, the method proposed in~\cite{huo2016object} uses color feature to detect salient objects. Some other methods use center prior to improve the performance of salient object detection~\cite{aksac2017complex, liang2018material}. Because of the lack of semantic information, these methods have limited ability to detect the whole structure of salient objects in complex scenes. In recent years, the methods based on the Fully Convolutional Neural Networks (FCNs), such as~\cite{luo2017non, zhang2017amulet, xi2019salient}, have been widely used for saliency detection owing to their high capacity of modeling high-level semantics. Even though these methods have achieved promising results, there are still some challenges due to the complicated scenarios of some images. The learned features by these methods usually lack the ability to i) suppress the non-salient regions that have "salient-like" appearance as depicted in the first row of Figure~\ref{fig:SM_first_page}, ii) detect salient objects that have different-looking regions as depicted in the second row of Figure~\ref{fig:SM_first_page}.

To address the above-mentioned challenges, we propose the Guide Module which takes advantage of the nature of the high-level and low-level features. By adopting this module, high-level features, which do not contain the fine spatial details of low-level features, can exploit the nature of low-level features as a guidance to make foreground and background regions more distinct, and thus it can suppress the non-salient regions that have "salient-like" appearance. For example, as illustrated in the first row of Figure~\ref{fig:SM_first_page}, although the triangular object has "salient-like" appearance, it should not be labeled as salient object, since it is not the most interesting and prominent part of the image. From Figure~\ref{fig:SM_first_page}, we can see that our method (denoted as GAGNet-R) is able to completely suppress the whole triangular object. Furthermore, by adopting the Guide Module, high-level features, which have the ability of category recognition of image regions because of containing high semantic information, can guide the selection of low-level features. By inspiring from the Channel Attention Block (CAB) proposed in~\cite{yu2018learning}, we give our model the ability to guide the selection of low-level features, which equips our network with the power of assigning foreground label to different-looking salient regions. As illustrated in the second row of Figure~\ref{fig:SM_first_page}, the appearance of the feet of the doll is different from the rest of the doll, but as it can be seen, our method is able to highlight the whole doll as the salient object. Thus, by benefiting from the content-aware guidance provided by our Guide Modules, our method is able to handle these complicated scenarios.

Some previous salient object detection methods~\cite{wang2016saliency, wang2015deep, zhang2017learning} utilize subsequent single-scale convolutional and max pooling layers to produce deep features. Since salient objects have large variations in scale and location, the learned features by these methods might not be able to handle these complicated variations due to the limited field of view. To extract multi-scale contextual information, some methods~\cite{zhang2018bi,kampffmeyer2018connnet} apply several parallel dilated convolutions with different rates inspired by structures such as Atrous Spatial Pyramid Pooling (ASPP)~\cite{chen2017deeplab}. However, the dilated convolution inserts "holes" in the convolution kernels to enlarge the receptive field, which would cause the loss of local information, especially when the dilation rate increases. This problem is called the "gridding issue" which was explored in~\cite{wang2018understanding}.
To address these problems, different from the methods based on dilated convolution and ASPP, we introduce the Multi-scale Feature Extraction Module (MFEM) which is capable of capturing multi-scale contextual information by enabling densely connections within the multi-scale regions in the feature map. For each level of abstraction (i.e., stage) of the pre-trained backbone, we perform convolutions by adopting a $3 \times 3$ trivial convolutional layer and Global Convolutional Networks (GCNs)~\cite{peng2017large} with different kernel sizes. Then, the resulting feature maps are stacked to form multi-scale features. GCNs enable densely connections within a large $k \times k$ region in the feature map and thus can alleviate the "gridding issue".

In this paper, we propose a Content-Aware Guidance Network, which we refer to as CAGNet, consisting of three networks: (i) Feature Extraction Network (FEN), (ii) Feature Guide Network (FGN), (iii) Feature Fusion Network (FFN). 

The FEN produces multi-scale features at multiple levels of abstraction by adopting the MFEM at each level of a pre-trained backbone. The FGN takes the extracted multi-scale features of the FEN as its input and guides the features in order to use by the FFN. Then, by using multiple add operations and Residual Refinement Modules (RRMs) in the FFN, the guided features are fused effectively. Our proposed RRM is a residual block with spatial attention, which refines the features with the ability of focusing on salient regions and avoiding distractions in the non-salient regions. In summary, the FEN, FGN, and FFN in our proposed architecture work collaboratively to generate more accurate saliency prediction. Additionally, while most saliency detection methods in the literature use the Cross-entropy loss to learn the salient objects, we design a loss function that outperforms the Cross-entropy by a large margin.

In this paper, by conducting experiments on various backbones, we prove the robustness of our method. Furthermore, our method contains a lower number of parameters in comparison with the previous state-of-the-art methods. It is worth mentioning that since salient object detection is a pre-processing step for many computer vision tasks, it is important to evaluate the performance in terms of the running speed. Our method is capable of running at a real-time speed of 28 FPS, which guarantees that our network can be practically adopted as a pre-processing step for computer vision tasks.

In short, our main contributions are summarized as follows: 
\begin{itemize}
\item We propose the Feature Guide Network to equip our model with the power of i) making the foreground and background regions more distinct and suppressing the non-salient regions that have "salient-like" appearance; ii) detecting salient objects that have different-looking regions. 
\item To extract powerful multi-scale features, different from the methods based on dilated convolution and ASPP, we propose a Multi-scale Feature Extraction Module which adopts GCNs to enable densely connections within large regions in the feature map. This module helps the model to alleviate the "gridding issue".
\item We design a loss function that outperforms the widely used Cross-entropy loss by a large margin.
\item Our method achieves great performance under different backbones, which shows that our proposed framework is very general with respect to the choice of the backbone model. It is interesting to note that while most methods in the saliency detection literature adopt a single backbone in their framework, we evaluate our framework on four different backbones to prove the generalization capability of our method.
\item The proposed method achieves the state-of-the-art on several challenging saliency detection datasets. Furthermore, our method contains a lower number of parameters compared to the previous state-of-the-art methods and can run at a real-time speed of 28 FPS.
\end{itemize}

\section{Related work}

\subsection{Video Salient Object Detection}

Video salient object detection has been widely applied in video segmentation, video compression, video captioning, weakly supervised attention, and autonomous driving. For example, Wang et al.~\cite{wang2017saliency} propose an unsupervised method that incorporates geodesic distance into saliency-aware video object segmentation. 

To detect salient object in videos, Wang et al.~\cite{wang2017video} propose a deep video saliency model which employs FCNs for pixel-wise prediction. This model is composed of two components, namely static saliency network and dynamic saliency network, which give the model the ability to capture spatial and temporal statistics of dynamic scenes. Shao et al.~\cite{guo2017video} introduce a method that identifies foreground regions in videos by using object proposals. This method detects salient objects by ranking and selecting the salient proposals based on object-level saliency cues. Fan et al.~\cite{fan2019shifting} propose a model for video salient object detection that is composed of two main components. The first component is Pyramid Dilated Convolution (PDC) which aims at robust static saliency representation learning. The second one is Saliency-Shift-aware convLSTM (SSLSTM) that is able to capture video saliency dynamics through modeling human visual attention-shift behavior. Lu et al.~\cite{lu2019see} propose a novel CO-attention Siamese Network (COSNet) which uses a co-attention mechanism to capture the temporal correlation across frames.

\subsection{Stereoscopic Image Salient Object Detection}

Some of the saliency detection methods are proposed to detect the salient regions in a stereoscopic three-dimensional (3D) image~\cite{niu2012leveraging, wang2016stereoscopic}. Niu et al.~\cite{niu2012leveraging} introduce a method for saliency detection, which is based on the global disparity contrast in a pair of stereo images.  Wang et al.~\cite{wang2016stereoscopic} propose a stereo saliency detection algorithm that considers stereoscopic information and the relevancy between the two views of a stereo pair. Different from these methods, in this paper, we focus on salient object detection in RGB Images.

\subsection{RGB Image Salient Object Detection}

Over the past years, numerous methods have been proposed for RGB image saliency detection. Traditional methods predict the saliency score based on hand-crafted features. Most of these methods utilize heuristic priors such as center prior~\cite{aksac2017complex, liang2018material}, boundary background~\cite{yang2013saliency}, and color contrast~\cite{cheng2014global}. Aytekin et al.~\cite{aytekin2018probabilistic} propose a probabilistic framework to encode the boundary connectivity saliency cue and smoothness constraints into a global optimization problem. Wang et al.~\cite{wang2016correspondence} propose a saliency transfer method to benefit from the existing large annotated datasets for recognizing the primary and smooth connected salient regions from an image. Shan et al.~\cite{shan2018visual} propose a graph-based approach and use background weight map to provide seeds for manifold ranking. Furthermore, they design a third-order smoothness framework to enhance the performance of manifold ranking.  These methods, which are based on the traditional approaches, fail to capture semantic and high-level information of the objects. 

Recently, deep Convolutional Neural Networks (CNNs) have shown their capabilities in extracting powerful features at multiple levels of abstraction. The CNN features can acquire a richer representation compared to the traditional hand-crafted features, and thus would result in performance improvement. In recent years, a vast number of methods have adopted CNNs for saliency detection task. For example, Li et al.~\cite{li2015visual} extract multi-scale features from a CNN and estimate the saliency score for each image super-pixel. Wang et al.~\cite{wang2015deep} employ two CNNs to combine local estimation of super-pixels and global proposal searching to predict saliency maps. Zhao et al.~\cite{zhao2015saliency} propose multi-context CNNs  for exploiting local and global context for salient object detection. Although these CNN-based methods have shown better performance than the traditional methods, they are time-consuming because of taking image patches as input. Moreover, these methods fail to consider important spatial information of the whole image. 

To overcome the above-mentioned problems, several methods have utilized FCNs to generate a pixel-wise prediction over the whole image directly. For instance, Li et al.~\cite{li2016deep} propose a multi-scale FCN  to explore the semantic properties and visual contrast information of salient objects. Hou et al.~\cite{hou2017deeply} introduce short connections to combine features in different layers. Zhang et al.~\cite{zhang2017amulet} propose a resolution-based feature combination module to integrate multi-level feature maps into multiple resolutions, which captures spatial details and semantic information simultaneously. Then, by fusing the predicted saliency maps in each resolution, the final saliency map is obtained. Zhang et al.~\cite{zhang2018bi} design a bi-directional message passing architecture to pass messages between multi-level features. Wang et al.~\cite{wang2018detect} propose to locate the salient objects globally and then refine them by taking advantage of local context information. Zhang et al.~\cite{zhang2019hyperfusion} employ a hyper-densely hierarchical feature fusion network to fuse the local and global multi-scale feature maps.

Most of the recent methods focus on using both high-level and low-level features for salient object detection. However, naively using these features may result in confusion for the network, and there needs to be an effective approach to use these features constructively. In this paper, we propose the Feature Guide Network which guides multi-level features to produce more effective features.

To obtain multi-scale features, some previous methods adopt parallel networks and feed them with re-scaled images~\cite{li2017instance} or multi-context super-pixels~\cite{li2015visual}. Different from these methods, we propose Multi-scale Feature Extraction Module (MFEM) to extract multi-scale features.

\section{Our method}
\begin{figure}[!t]
\begin{center}
\includegraphics[width=0.9\textwidth]{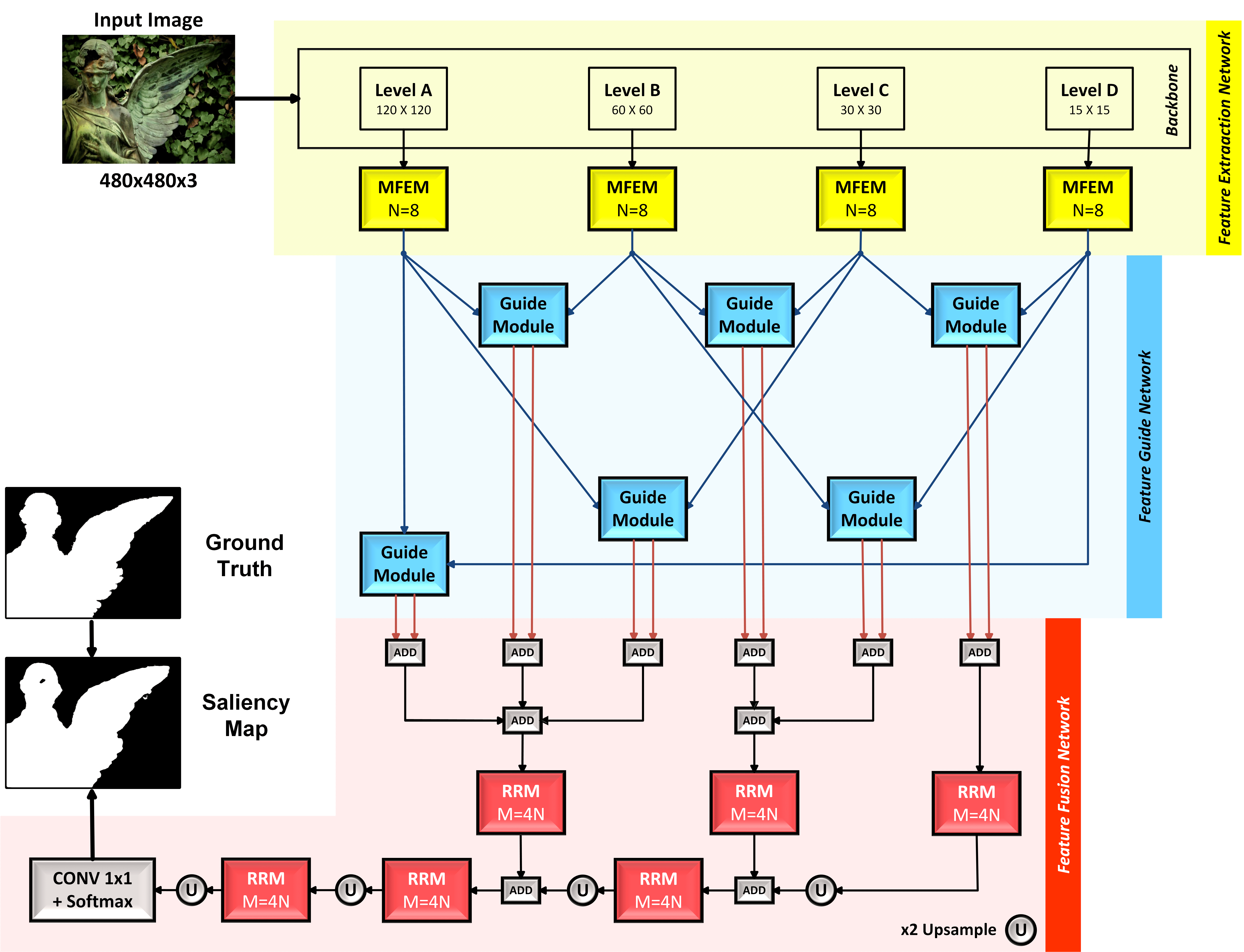}
\end{center}
  \caption{The overall architecture of our proposed Content-Aware Guidance Network (CAGNet). CAGNet consists of three networks: (i) Feature Extraction Network which captures multi-scale contextual features, (ii) Feature Guide Network which guides the extracted features by taking advantage of the nature of high-level and low-level features, (iii) Feature Fusion Network which fuses the guided features effectively to generate the saliency map.}
\label{fig:Main}
\end{figure}

In this section, we first explain our proposed Content-Aware Guidance Network (CAGNet) containing three networks: (i) Feature Extraction Network which extracts multi-scale context information, (ii) Feature Guide Network which guides the extracted features by taking advantage of the spatial details of low-level features and the semantic information of high-level features, (iii) Feature Fusion Network which integrates the guided features effectively to generate the saliency map. The architecture of the proposed CAGNet is illustrated in Figure~\ref{fig:Main}. Finally, we describe our designed loss function that has better performance than the widely used Cross-entropy loss.

\subsection{Feature Extraction Network}
Feature Extraction Network consists of a pre-trained backbone that takes the input image and produces multi-level feature maps, and Multi-scale Feature Extraction Modules (MFEMs) which we apply them to multi-level feature maps to capture multi-scale contextual features. 

\subsubsection{Pre-trained backbone}
In this study, we examine different pre-trained models in our CAGNet as the backbone model, namely VGG-16~\cite{simonyan2014very}, ResNet50~\cite{he2016deep}, NASNet-Mobile~\cite{zoph2018learning}, and NASNet-large~\cite{zoph2018learning}, which are denoted as CAGNet-V, CAGNet-R, CAGNet-M, and CAGNet-L, respectively. These backbones are used to produce features at different levels of abstraction. To fit the need of saliency detection task, we remove all fully connected layers in these backbones. In VGG-16, the features after the last max pooling layer cannot introduce a new level of abstraction. Thus, we use a convolutional layer with $1024$ kernels of size $3 \times 3$ after the last max pooling layer in VGG-16 to produce a new level.

The output feature maps of all backbones are re-scaled by a factor of $32$ with respect to the input image. We take feature maps at four levels from each backbone. Given an input image with size $W \times H$, these feature maps have spatial sizes of $W/2^n \times H/2^n$ with $n=2,3,4,5$. The details of selected layers for different levels of abstraction in each backbone are shown in Table~\ref{tab:Levels}.

\begin{table}
\centering
\caption{Selected layers for different levels of abstraction in the adopted backbones. Note that we take the output of these layers for feature extraction. The size of the feature maps are shown in parentheses. The Level D in CAGNet-V (which is denoted as The added layer) is obtained by adding $1024$ kernels of size $3\times 3$ after the last max pooling layer of the original VGG-16.}
\label{tab:Levels}
\resizebox{0.7\linewidth}{!}{%
{\renewcommand{\arraystretch}{1.4}
\begin{tabular}{@{}l|cccc@{}}
\toprule
Backbone         & Level A                                                                      & Level B                                                                  & Level C                                                                   & Level D                                                                         \\ \toprule
VGG-16~\cite{simonyan2014very}        & \begin{tabular}[c]{@{}c@{}}Conv3-3\\ ($120\times 120\times 256$)\end{tabular}    & \begin{tabular}[c]{@{}c@{}}Conv4-3\\ ($60\times 60\times 512$)\end{tabular}  & \begin{tabular}[c]{@{}c@{}}Conv5-3\\ ($30\times 30\times 512$)\end{tabular}   & \begin{tabular}[c]{@{}c@{}}The added layer\\ ($15\times 15\times 1024$)\end{tabular} \\ \midrule
ResNet50~\cite{he2016deep}      & \begin{tabular}[c]{@{}c@{}}Conv2-x\\ ($120\times 120\times 256$)\end{tabular}    & \begin{tabular}[c]{@{}c@{}}Conv3-x\\ ($60\times 60\times 512$)\end{tabular}  & \begin{tabular}[c]{@{}c@{}}Conv4-x\\ ($30\times 30\times 1024$)\end{tabular}  & \begin{tabular}[c]{@{}c@{}}Conv5-x\\ ($15\times 15\times 2048$)\end{tabular}      \\ \midrule
NASNet-Mobile~\cite{zoph2018learning} & \begin{tabular}[c]{@{}c@{}}1st reduction cell\\ ($120\times 120\times 44$)\end{tabular}  & \begin{tabular}[c]{@{}c@{}}4th normal cell\\ ($60\times 60\times 264$)\end{tabular} & \begin{tabular}[c]{@{}c@{}}8th normal cell\\ ($30\times 30\times 528$)\end{tabular}   & \begin{tabular}[c]{@{}c@{}}12th normal cell\\ ($15\times 15\times 1056$)\end{tabular}  \\ \midrule
NASNet-Large~\cite{zoph2018learning}  & \begin{tabular}[c]{@{}c@{}}1st reduction cell\\ ($120\times 120\times 168$)\end{tabular} & \begin{tabular}[c]{@{}c@{}}6th normal cell\\ ($60\times 60\times 1008$)\end{tabular} & \begin{tabular}[c]{@{}c@{}}12th normal cell\\ ($30\times 30\times 2016$)\end{tabular} & \begin{tabular}[c]{@{}c@{}}18th normal cell\\ ($15\times 15\times 4032$)\end{tabular}   \\ \bottomrule
\end{tabular}%
}
}
\end{table}

\subsubsection{Multi-scale Feature Extraction Module}
Salient objects have large variations in scale and location in different images. Due to the variability of scale, using single scale convolution may not capture the right size. Moreover, due to the variability of location, using pyramid pooling as a multi-scale feature extractor, as proposed in~\cite{wang2017stagewise}, would cause the loss of important local information because of the large scale of pooling. Another approach to implement a multi-scale feature extractor is to use dilated convolutions like~\cite{zhang2018bi}, which enlarges the receptive field by inserting "holes" in the convolution kernels, and thus would result in the loss of local information because of sparse connections. This problem, which is called the "gridding issue", was explored in~\cite{wang2018understanding}.

Based on above observation, we find the Global Convolutional Networks (GCNs)~\cite{peng2017large} effective to address the "gridding issue" challenge. To avoid sparse connections and enable densely connections within a large $k \times k$ region in the feature map, GCN utilizes a combination of $k \times 1 + 1 \times k$ and $1 \times k + k \times 1$ convolutions to implement the $k \times k$ convolution effectively with a lower number of parameters compared to the trivial $k \times k$ convolution. More details about the GCN can be found in~\cite{peng2017large}. Furthermore, to obtain multi-scale contextual information, by taking advantage of GCNs, we propose the Multi-scale Feature Extraction Module (MFEM). This module contains GCNs with different kernel sizes and can learn multi-scale context information at multiple levels of abstraction. 

\begin{figure}[tp]
\begin{center}
\includegraphics[width=0.6\linewidth]{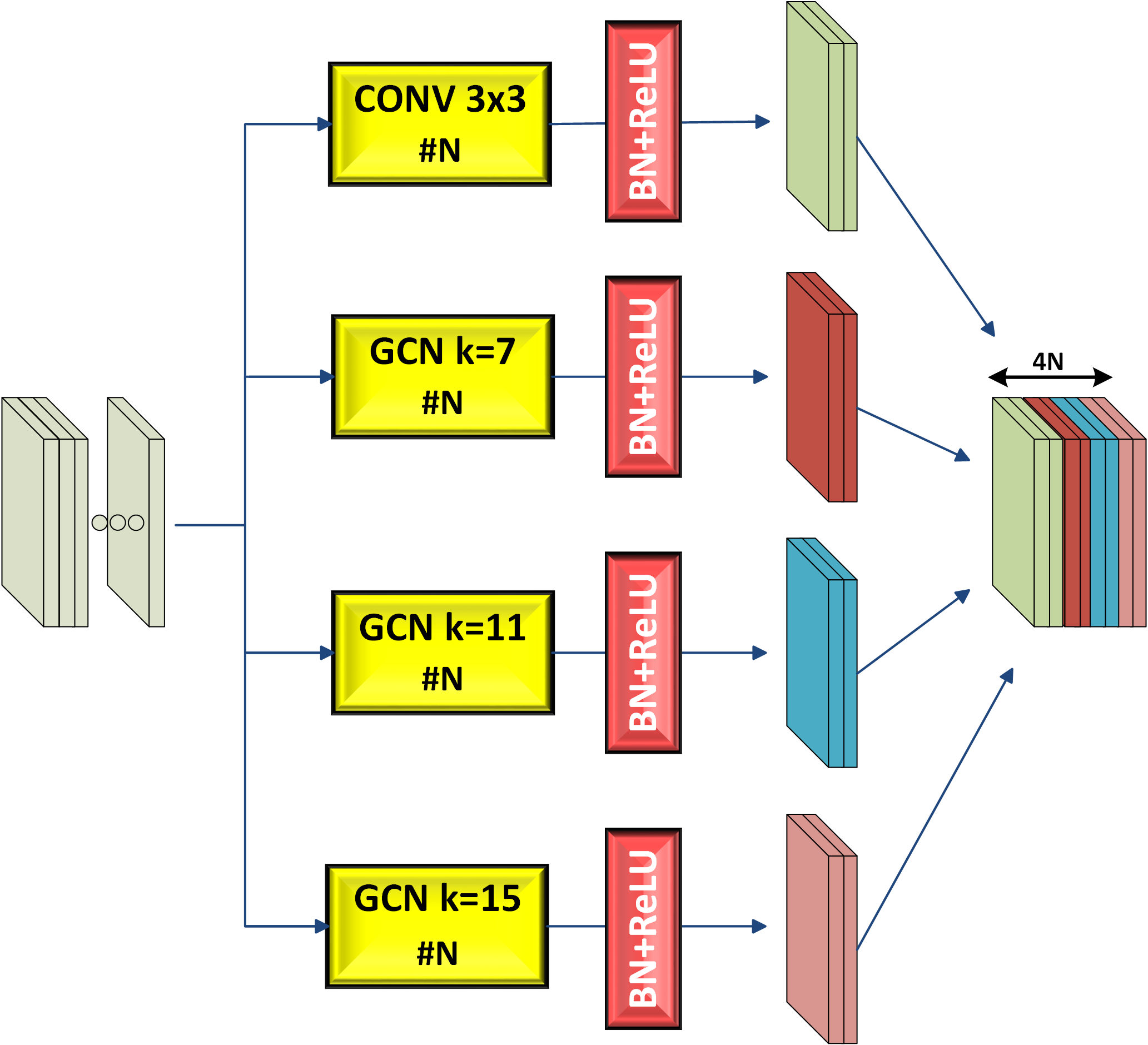}
\end{center}
  \caption{Multi-scale Feature Extraction Module (MFEM). MFEM adopts the $3 \times 3$ trivial convolution and GCNs with $k = 7, 11, 15$ to extract multi-scale features. The '$ \# $' symbol denotes the number of layer filters. This figure shows MFEM with N=2.
}
\label{fig:MFEM}
\end{figure}

As illustrated in Figure~\ref{fig:MFEM}, in the MFEM, we perform convolutions by utilizing  the $3 \times 3$ trivial convolution and GCNs with $k=7,11,15$. Then, the resulting feature maps are concatenated to form multi-scale features. 

\subsection{Feature Guide Network}
We employ four MFEMs in the Feature Extraction Network to extract multi-scale features at four different levels of abstraction. These different levels have different recognition information. High-level features have semantic and global information because of the large field of view. Thus, these features can help the category recognition of image regions. Low-level features have spatial and local information due to the small field of view. Therefore, the information of low-level features can help to better locate the salient regions. 

\begin{figure}[tp]
\begin{center}
\includegraphics[width=0.8\linewidth]{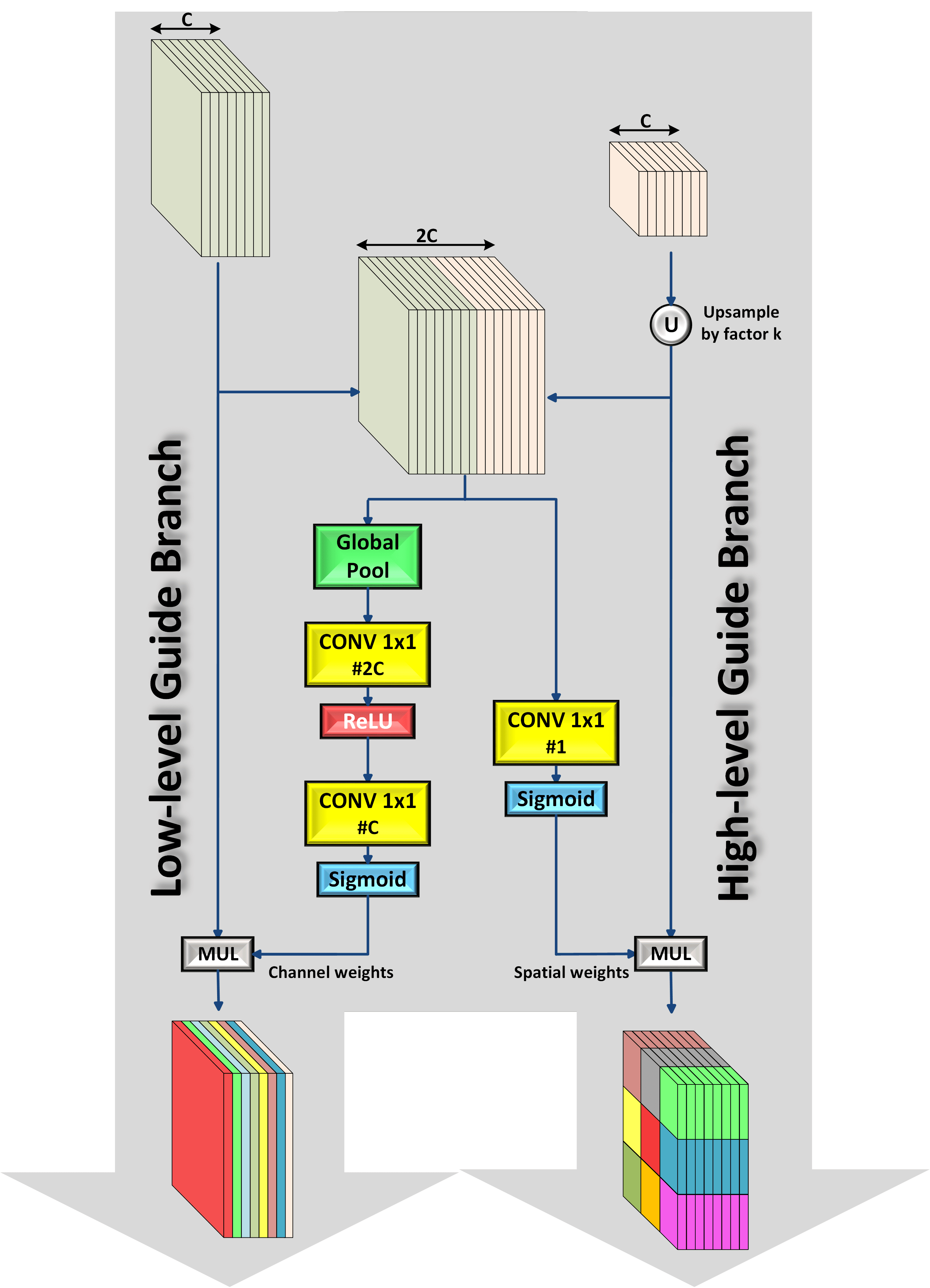}
\end{center}
  \caption{The illustration of the Guide Module. This module consists of High-level Guide and Low-level Guide branches and is adopted to guide the features of the different levels. Note that C shows the number of the channels of the input feature maps, and the '$ \# $' symbol denotes the number of layer filters.
}
\label{fig:Guide}
\end{figure}

Based on above observation, we propose the Feature Guide Network to better exploit the diverse recognition abilities of different levels. Feature Guide Network is composed of multiple Guide Modules which help to produce more powerful features for saliency detection. As illustrated in Figure~\ref{fig:Guide}, Guide Module consists of Low-level Guide and High-level Guide branches. This module takes low-level and high-level features as its inputs and outputs guided low-level and guided high-level features.

In saliency detection, some non-salient regions may have "salient-like" appearance. As shown in the first row of Figure~\ref{fig:SM_first_page}, the triangular object at the bottom of the image, which has "salient-like" appearance, may cause confusion for saliency prediction. To address this challenge, we take advantage of the nature of the lower levels to guide the higher levels. In the lower levels, the Feature Extraction Network captures finer spatial information because of its smaller field of view compared to the higher levels. Thus, by applying a $1 \times 1$ convolution on concatenated high-level and low-level features, spatial weights are produced to weight the spatial information of high-level features. With this design, high-level features, which lack the low-level cues, can exploit the fine spatial details of low-level features as a guidance to make salient and non-salient regions more distinguishable. Therefore, by guiding the spatial information of high-level features, our network is able to enhance the distinction of salient and non-salient regions and suppress the non-salient regions with "salient-like" appearance.

In some complicated scenarios, salient regions may have different appearances. As illustrated in the second row of Figure~\ref{fig:SM_first_page}, the appearance of the feet of the doll is different from the rest of the doll. Assigning foreground label to these different-looking regions is challenging. To address this challenge, by inspiring from the Channel Attention Block (CAB) proposed in~\cite{yu2018learning}, we use the nature of high-level features to guide low-level features in our Feature Guide Network. High-level features have higher semantic information due to the large receptive field. By applying an architecture like Squeeze and Excitation Networks~\cite{hu2018squeeze} on concatenated high-level and low-level features, channel weights are generated to weight the channels of low-level feature maps. In this way, by utilizing high-level semantic information, the low-level features are guided to produce more attentive features. Thus, Guide Modules provide content-aware guidance for multi-level features, which would result in more accurate saliency prediction.

\subsection{Feature Fusion Network}
\begin{figure}
\begin{center}
\includegraphics[width=0.25\linewidth]{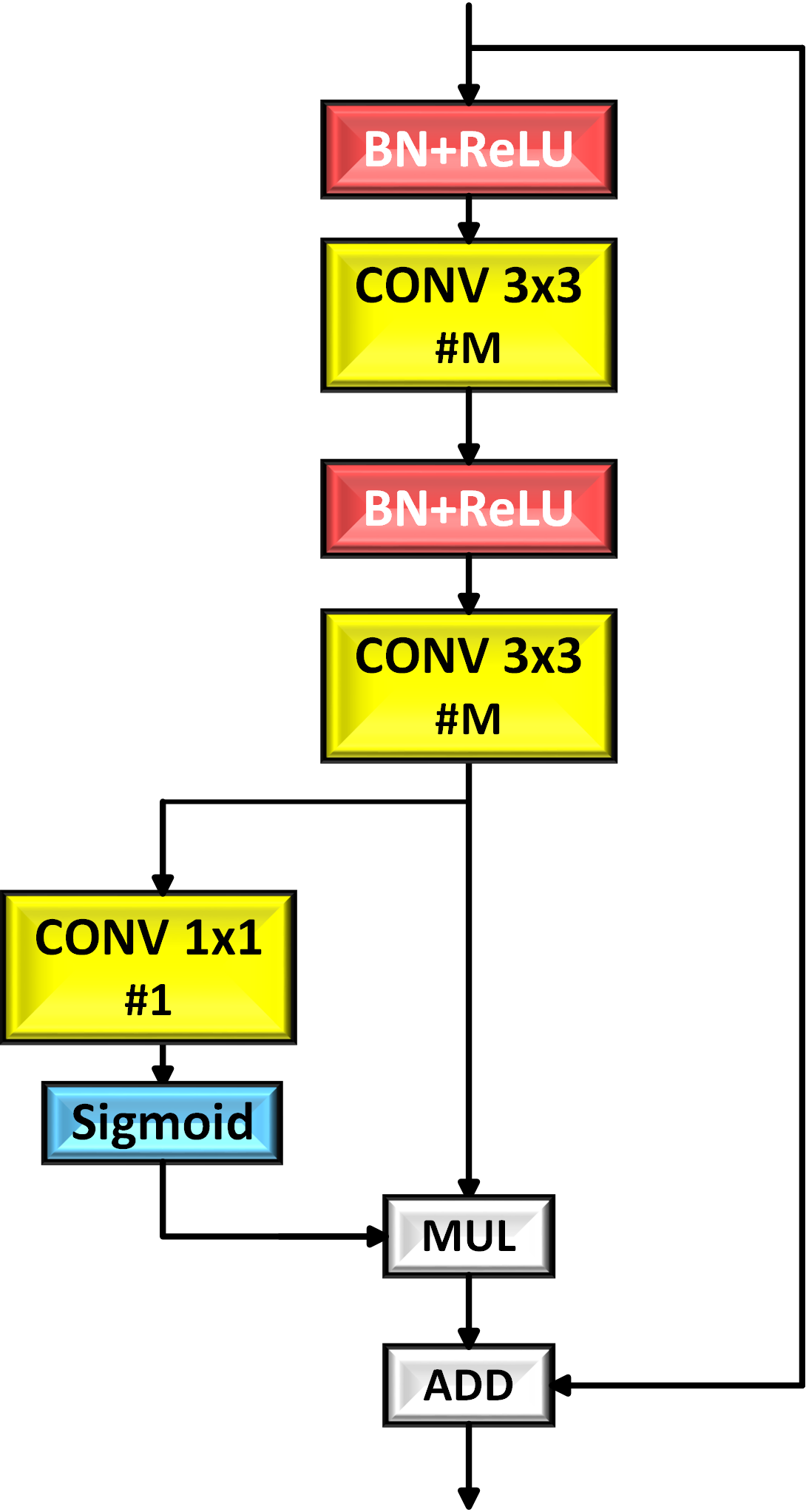}
\end{center}
  \caption{Residual Refinement Module (RRM). RRM is a residual block with spatial attention and is adopted to refine the features effectively. The '$ \# $' symbol denotes the number of layer filters.}
\label{fig:RRM}
\end{figure}

By adopting Feature Extraction Network and Feature Guide Network, guided multi-scale features at different levels of abstractions are obtained. To integrate these features effectively, we propose the Feature Fusion Network. In this network, we use add operations to combine different feature maps. In order to refine the features effectively, we introduce Residual Refinement Module (RRM), which is schematically depicted in Figure~\ref{fig:RRM}. RRM is a residual block~\cite{he2016deep, he2016identity} with spatial attention. This module is used to refine the features and has the ability of focusing on salient regions and avoiding distractions in the non-salient regions. Several works such as~\cite{yu2018learning} and~\cite{peng2017large} have used the standard residual block as their refinement module. Following their works, we have used the full pre-activation version of the residual block along with a spatial attention branch.

By adopting multiple RRM modules and add operations in Feature Fusion Network, finally the saliency map is obtained by utilizing a convolutional layer with two $1\times 1$ kernels with softmax activation.

\subsection{Our designed loss function for learning the salient objects}
In saliency detection literature, Cross-entropy loss function is widely used for learning the salient objects. However, the networks trained with Cross-entropy loss often differentiate boundary pixels with low confidence, which would result in performance degradation. In this paper, we design a loss function that leads to better results compared to the Cross-entropy loss, as shown in the ablation analysis section. Let  $I=\{I_m , m= 1,$ ...$,M\}$, $S_m$, and $G_m$ denote the training images, saliency map for the $m$-$th$ training image, and ground truth for the $m$-$th$ training image, respectively. Our designed loss is formulated as: 

\begin{equation}
L=\alpha_1 L_P + \alpha_2 L_R + \alpha_3L_{MAE}
\end{equation}
\noindent where $\alpha_1$ , $\alpha_2$ and $\alpha_3$ are the balance parameters. We empirically set $\alpha_1  = 1 $, $\alpha_2 = 0.5$, and $\alpha_3 = 1$. $L_P$ and $L_R$ are computed as:

\begin{equation}
L_P=1-\frac{1}{M}\sum_{m=1}^M P{(S_m, G_m)}
\end{equation}

\begin{equation}
L_R=1-\frac{1}{M}\sum_{m=1}^M R{(S_m, G_m)}
\end{equation}
\noindent where  $P(S.G)$ and $R(S,G)$ are calculated similar to Precision and Recall:

\begin{equation}
P(S,G) = \frac{\sum_n s_ng_n}{\sum_n s_n+\epsilon}
\end{equation}

\begin{equation}
 R(S,G)=\frac{\sum_n s_ng_n}{\sum_n g_n+\epsilon}
\end{equation}
\noindent where $s_n \in S$ and $g_n \in G$ , and $\epsilon$ is a regularization constant.  $L_{MAE}$ calculates the discrepancy between the predicted saliency map $S$ and the ground truth $G$:

\begin{equation}
L_{MAE} = \frac{1}{M}\sum_{m=1}^M MAE(S_m,G_m)
\end{equation}
\noindent where $MAE(S,G)$ is computed as :
\begin{equation}
MAE(S,G) =\frac{1}{N}\sum_{n} \mid s_n - g_n \mid
\end{equation}
\noindent where $N$ denotes the total number of pixels. In ablation analysis section, we demonstrate that our designed loss function outperforms  the Cross-entropy loss function.

\section{Experiments}
\subsection{Datasets and evaluation metrics}

The proposed method is evaluated on six public salient object detection datasets.
ECSSD~\cite{yan2013hierarchical} contains 1,000 semantically meaningful and complex images with multiple objects of different sizes. 
DUT-OMRON~\cite{yang2013saliency} consists of 5,168 challenging images with high variety of content, each of which has complex background and one or two salient objects. 
HKU-IS~\cite{li2015visual} contains 4447 images with low color contrast. Images in this dataset are selected to include multiple foreground objects or objects touching the image boundary. 
DUTS~\cite{wang2017learning} dataset is currently the largest salient object detection dataset and comprised of 10,553 images in the training set and 5,019 images in the test set. Both training and test sets have very challenging scenarios. 
The PASCAL-S~\cite{li2014secrets} dataset has 850 natural images chosen from the PASCAL VOC 2010~\cite{everingham2010pascal} segmentation dataset. In SOC~\cite{fan2018salient} dataset, each salient image is accompanied by attributes that reflect common challenges in real-world scenarios. We use the validation set of this dataset for testing.

We use six metrics to evaluate the performance of our method as well as previous state-of-the-art saliency detection methods, namely Precision-Recall (PR) curves, F-measure curves, Average F-measure (denoted as avgF) score, weighted F-measure (denoted as wF) score, E-measure (denoted as E) score, and Mean Absolute Error (denoted as MAE) score. More detailed descriptions about these metrics can be found in~\cite{borji2015salient, margolin2014evaluate, fan2018enhanced}. Furthermore, the code for computing these metrics in Python can be found at {\href{https://github.com/Mehrdad-Noori/Saliency-Evaluation-Toolbox}{https://github.com/Mehrdad-Noori/Saliency-Evaluation-Toolbox}}.

Precision is the fraction of correct salient pixels in the predicted saliency map, and Recall is defined as the fraction of correct salient pixels in the ground truth. To calculate Precision and Recall, the binarized saliency map is compared against the ground truth mask. The threshold is varied from 0 to 1 to generate a sequence of binary masks. These binary masks are used to calculate (Precision, Recall) pairs and (F-measure, threshold) pairs to plot the PR curves and the F-measure curves.

The Average F-measure score is calculated by using the thresholding method suggested in~\cite{achanta2009frequency}. This threshold is used to generate binary maps for computing the F-measure which is defined as:

\begin{equation}
F_\beta=\frac{(1+\beta^2)\cdot Precision\cdot Recall}{\beta^2\cdot Precision + Recall}
\end{equation}

\noindent where $\beta^2$ is set to 0.3 to weight precision more than recall.
The weighted F-measure score~\cite{margolin2014evaluate} and E-measure score~\cite{fan2018enhanced} are also adopted to evaluate the performance. Finally, the MAE score is calculated as the average pixel-wise absolute difference between the ground truth mask and the predicted saliency map.

\subsection{Implementation details}
We develop our proposed method in Keras~\cite{chollet2015keras}  framework using TensorFlow~\cite{tensorflow2015-whitepaper} backend. The backbone models (i.e., VGG-16~\cite{simonyan2014very}, ResNet-50~\cite{he2016deep}, NASNet Mobile~\cite{zoph2018learning}, and NASNet Large~\cite{zoph2018learning}) are initialized with ImageNet~\cite{russakovsky2015imagenet} weights. In our experiments, the input image is resized into $480\times 480$ pixels for training and testing. To reduce overfitting, two types of data augmentation are randomly employed: horizontal filliping and rotation (range of 0-12 degrees). We do not use validation set and train the model until its training loss converges. All the experiments are performed using the stochastic gradient descent with a momentum coefficient $0.9$ and an initial learning rate of $8e$-$3$ which is divided by $10$ if no improvement in training loss is seen for $10$ epochs. We perform our experiments on an NVIDIA 1080 Ti GPU. The code, the trained models, and the saliency maps of our method can be found at {\href{https://github.com/Mehrdad-Noori/CAGNet}{https://github.com/Mehrdad-Noori/CAGNet}}.

\subsection{Comparison with the state-of-the-art}
We compare our method with 16 previous state-of-the-art methods, namely MDF~\cite{li2015visual}, RFCN~\cite{wang2016saliency}, UCF~\cite{zhang2017learning}, Amulet~\cite{zhang2017amulet}, NLDF~\cite{luo2017non}, DSS~\cite{hou2017deeply}, BMPM~\cite{zhang2018bi}, PAGR~\cite{zhang2018progressive}, PiCANet~\cite{liu2018picanet}, SRM~\cite{wang2017stagewise}, DGRL~\cite{wang2018detect}, MLMS~\cite{wu2019mutual}, AFNet~\cite{feng2019attentive}, CapSal~\cite{zhang2019capsal}, BASNet~\cite{qin2019basnet}, and CPD~\cite{wu2019cascaded}. For a fair comparison, we use the saliency maps provided by the authors.

\begin{figure}[!t]
\begin{center}
\includegraphics[width=0.95\textwidth]{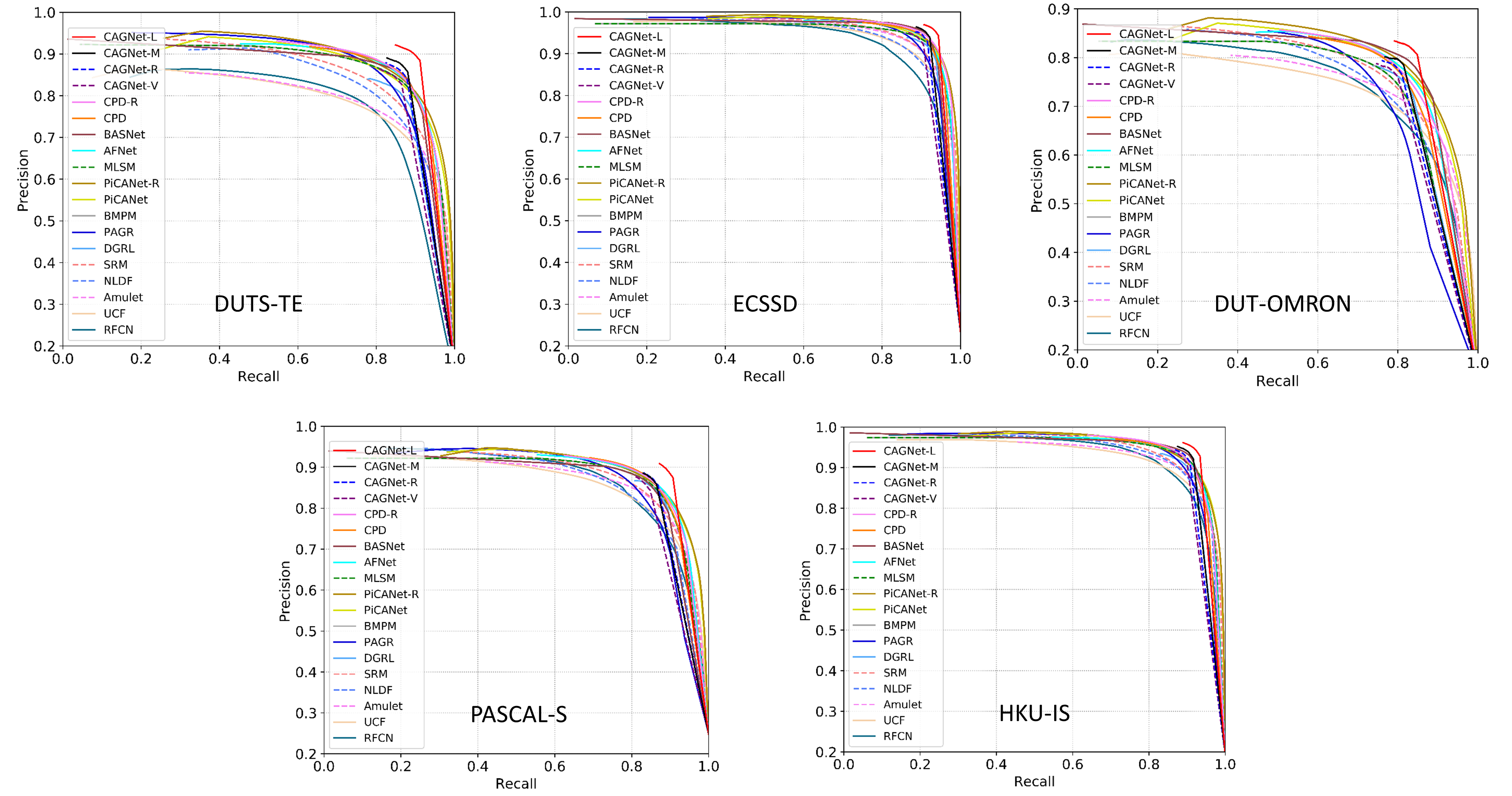}
\end{center}
  \caption{The PR curves of the proposed method and previous state-of-the-art methods.}
\label{fig:PRcurves}
\end{figure}

\begin{figure}[!t]
\begin{center}
\includegraphics[width=0.95\textwidth]{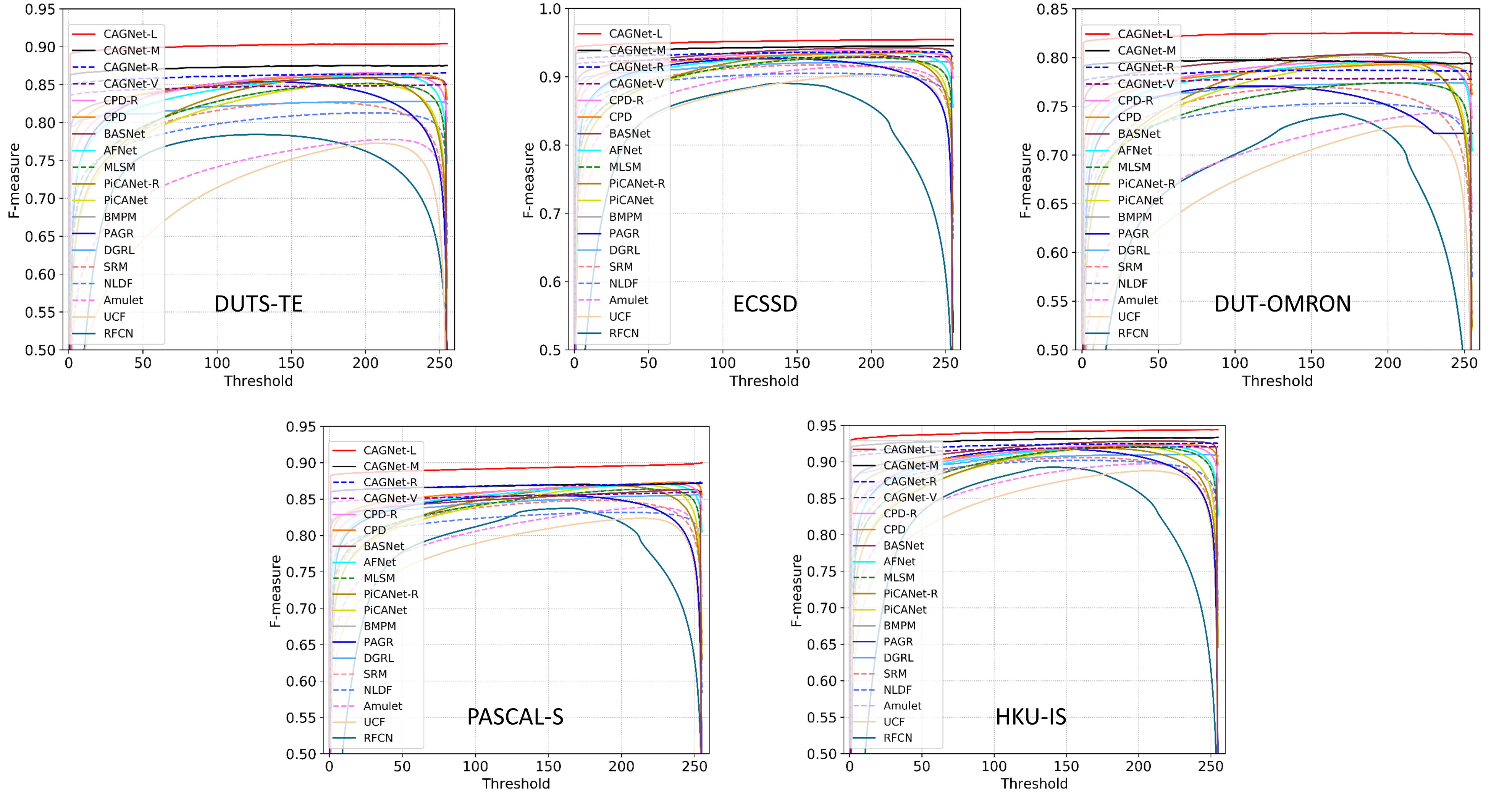}
\end{center}
  \caption{The F-measure curves of the proposed method and previous state-of-the-art methods.}
\label{fig:Fcurves}
\end{figure}

\begin{table}[!t]
\centering
\caption{Comparison of the proposed method and other 16 methods on five salient object detection datasets in terms of avgF, wF, E-measure, and MAE scores. CAGNet with VGG-16, ResNet50, NASNet-Mobile, and NASNet-Large backbones, are denoted as CAGNet-V, CAGNet-R, CAGNet-M, and CAGNet-L, respectively. The best score and the second best score under each setting are shown in {\color[HTML]{FE0000}\textbf{red}} and {\color[HTML]{3166FF}\textbf{blue}}, respectively, and the best score under all settings is underlined. The unit of the total number of parameters (denoted as \#Par) is million. Note that the authors of ~\cite{zhang2018progressive} did not release the code, and they just provided the saliency maps, and thus reporting the total number of parameters of this method is not possible.}
\label{tab:comparison}
\resizebox{\textwidth}{!}{%
{\renewcommand{\arraystretch}{1.6}
\setlength\tabcolsep{2pt}
\begin{tabular}{lccccccccccccccccccccccc}
\hline
\multicolumn{1}{l|}{Dataset}           & \multicolumn{1}{c|}{}                       & \multicolumn{1}{c|}{}                           & \multicolumn{1}{c|}{}                                           & \multicolumn{4}{c|}{DUTS-TE~\cite{wang2017learning}}                                                                                                                                                                               & \multicolumn{4}{c|}{ECSSD~\cite{yan2013hierarchical}}                                                                                                                                                                                 & \multicolumn{4}{c|}{DUT-O~\cite{yang2013saliency}}                                                                                                                                                                                 & \multicolumn{4}{c|}{PASCAL-S~\cite{li2014secrets}}                                                                                                                                                                              & \multicolumn{4}{c}{HKU-IS~\cite{li2015visual}}                                                                                                                                                           \\ \cline{1-1} \cline{5-24} 
\multicolumn{1}{l|}{Metric}            & \multicolumn{1}{c|}{\multirow{-2}{*}{Year}} & \multicolumn{1}{c|}{\multirow{-2}{*}{Backbone}} & \multicolumn{1}{c|}{\multirow{-2}{*}{\#Par}}                    & avgF                                        & wF                                          & E                                           & \multicolumn{1}{c|}{MAE}                                         & avgF                                        & wF                                          & E                                           & \multicolumn{1}{c|}{MAE}                                         & avgF                                        & wF                                          & E                                           & \multicolumn{1}{c|}{MAE}                                         & avgF                                        & wF                                          & E                                           & \multicolumn{1}{c|}{MAE}                                         & avgF                                        & wF                                          & E                                           & MAE                                         \\ \hline 
\multicolumn{24}{c}{VGG~\cite{simonyan2014very}}                                                                                                                                                                                                                                                                                                                                                                                                                                                                                                                                                                                                                                                                                                                                                                                                                                                                                                                                                                                                                                                                                                                                                                                                                      \\ \hline
\multicolumn{1}{l|}{MDF~\cite{li2015visual}}               & \multicolumn{1}{c|}{2015}                   & \multicolumn{1}{c|}{VGG16}                      & \multicolumn{1}{c|}{56.86}                                      & 0.669                                       & 0.588                                       & 0.813                                       & \multicolumn{1}{c|}{0.093}                                       & 0.807                                       & 0.705                                       & 0.853                                       & \multicolumn{1}{c|}{0.105}                                       & 0644                                        & 0.564                                       & 0.802                                       & \multicolumn{1}{c|}{0.092}                                       & 0.711                                       & 0.590                                       & 0.759                                       & \multicolumn{1}{c|}{0.146}                                       & 0.784                                       & 0.564                                       & 0.871                                       & 0.129                                       \\
\multicolumn{1}{l|}{RFCN~\cite{wang2016saliency}}              & \multicolumn{1}{c|}{2016}                   & \multicolumn{1}{c|}{VGG16}                      & \multicolumn{1}{c|}{134.69}                                     & 0.711                                       & 0.586                                       & 0.840                                       & \multicolumn{1}{c|}{0.090}                                       & 0.834                                       & 0.698                                       & 0.877                                       & \multicolumn{1}{c|}{0.107}                                       & 0.627                                       & 0.524                                       & 0.779                                       & \multicolumn{1}{c|}{0.110}                                       & 0.754                                       & 0.636                                       & 0.810                                       & \multicolumn{1}{c|}{0.132}                                       & 0.835                                       & 0.680                                       & 0.906                                       & 0.089                                       \\
\multicolumn{1}{l|}{UCF~\cite{zhang2017learning}}               & \multicolumn{1}{c|}{2017}                   & \multicolumn{1}{c|}{VGG16}                      & \multicolumn{1}{c|}{23.98}                                      & 0.631                                       & 0.596                                       & 0.770                                       & \multicolumn{1}{c|}{0.112}                                       & 0.844                                       & 0.806                                       & 0.895                                       & \multicolumn{1}{c|}{0.069}                                       & 0.621                                       & 0.573                                       & 0.768                                       & \multicolumn{1}{c|}{0.120}                                       & 0.738                                       & 0.700                                       & 0.809                                       & \multicolumn{1}{c|}{0.116}                                       & 0.823                                       & 0.779                                       & 0.904                                       & 0.062                                       \\
\multicolumn{1}{l|}{Amulet~\cite{zhang2017amulet}}            & \multicolumn{1}{c|}{2017}                   & \multicolumn{1}{c|}{VGG16}                      & \multicolumn{1}{c|}{33.15}                                      & 0.678                                       & 0.658                                       & 0.803                                       & \multicolumn{1}{c|}{0.085}                                       & 0.868                                       & 0.840                                       & 0.912                                       & \multicolumn{1}{c|}{0.059}                                       & 0.647                                       & 0.626                                       & 0.784                                       & \multicolumn{1}{c|}{0.098}                                       & 0.771                                       & 0.741                                       & 0.831                                       & \multicolumn{1}{c|}{0.099}                                       & 0.841                                       & 0.817                                       & 0.914                                       & 0.051                                       \\
\multicolumn{1}{l|}{NLDF~\cite{luo2017non}}              & \multicolumn{1}{c|}{2017}                   & \multicolumn{1}{c|}{VGG16}                      & \multicolumn{1}{c|}{35.49}                                      & 0.739                                       & 0.710                                       & 0.855                                       & \multicolumn{1}{c|}{0.065}                                       & 0.878                                       & 0.839                                       & 0.912                                       & \multicolumn{1}{c|}{0.063}                                       & 0.684                                       & 0.634                                       & 0.817                                       & \multicolumn{1}{c|}{0.080}                                       & 0.782                                       & 0.742                                       & 0.842                                       & \multicolumn{1}{c|}{0.101}                                       & 0.873                                       & 0.838                                       & 0.929                                       & 0.048                                       \\
\multicolumn{1}{l|}{DSS~\cite{hou2017deeply}}               & \multicolumn{1}{c|}{2017}                   & \multicolumn{1}{c|}{VGG16}                      & \multicolumn{1}{c|}{62.23}                                      & 0.716                                       & 0.702                                       & 0.845                                       & \multicolumn{1}{c|}{0.065}                                       & 0.873                                       & 0.836                                       & 0.915                                       & \multicolumn{1}{c|}{0.062}                                       & 0.674                                       & 0.643                                       & 0.820                                       & \multicolumn{1}{c|}{0.074}                                       & 0.776                                       & 0.728                                       & 0.836                                       & \multicolumn{1}{c|}{0.103}                                       & 0.856                                       & 0.821                                       & 0.926                                       & 0.050                                       \\
\multicolumn{1}{l|}{PAGR~\cite{zhang2018progressive}}              & \multicolumn{1}{c|}{2018}                   & \multicolumn{1}{c|}{VGG19}                      & \multicolumn{1}{c|}{---}                                        & 0.784                                       & 0.724                                       & 0.883                                       & \multicolumn{1}{c|}{0.055}                                       & 0.894                                       & 0.833                                       & 0.917                                       & \multicolumn{1}{c|}{0.061}                                       & 0.711                                       & 0.622                                       & 0.843                                       & \multicolumn{1}{c|}{0.071}                                       & 0.808                                       & 0.738                                       & 0.854                                       & \multicolumn{1}{c|}{0.095}                                       & 0.886                                       & 0.820                                       & 0.939                                       & 0.047                                       \\
\multicolumn{1}{l|}{BMPM~\cite{zhang2018bi}}              & \multicolumn{1}{c|}{2018}                   & \multicolumn{1}{c|}{VGG16}                      & \multicolumn{1}{c|}{22.09}                                      & 0.745                                       & 0.761                                       & 0.863                                       & \multicolumn{1}{c|}{0.049}                                       & 0.868                                       & 0.871                                       & 0.916                                       & \multicolumn{1}{c|}{0.045}                                       & 0.692                                       & 0.681                                       & 0.839                                       & \multicolumn{1}{c|}{0.064}                                       & 0.771                                       & 0.785                                       & 0.847                                       & \multicolumn{1}{c|}{0.075}                                       & 0.871                                       & 0.859                                       & 0.938                                       & 0.039                                       \\
\multicolumn{1}{l|}{PiCANet~\cite{liu2018picanet}}           & \multicolumn{1}{c|}{2018}                   & \multicolumn{1}{c|}{VGG16}                      & \multicolumn{1}{c|}{32.85}                                      & 0.749                                       & 0.747                                       & 0.865                                       & \multicolumn{1}{c|}{0.054}                                       & 0.885                                       & 0.865                                       & 0.926                                       & \multicolumn{1}{c|}{0.046}                                       & 0.710                                       & 0.691                                       & 0.842                                       & \multicolumn{1}{c|}{0.068}                                       & 0.804                                       & 0.781                                       & 0.862                                       & \multicolumn{1}{c|}{0.079}                                       & 0.870                                       & 0.847                                       & 0.938                                       & 0.042                                       \\
\multicolumn{1}{l|}{MLMS~\cite{wu2019mutual}}              & \multicolumn{1}{c|}{2019}                   & \multicolumn{1}{c|}{VGG16}                      & \multicolumn{1}{c|}{74.38}                                      & 0.745                                       & 0.761                                       & 0.863                                       & \multicolumn{1}{c|}{0.049}                                       & 0.868                                       & 0.871                                       & 0.916                                       & \multicolumn{1}{c|}{0.044}                                       & 0.692                                       & 0.681                                       & 0.839                                       & \multicolumn{1}{c|}{0.064}                                       & 0.771                                       & 0.785                                       & 0.847                                       & \multicolumn{1}{c|}{0.075}                                       & 0.871                                       & 0.859                                       & 0.938                                       & 0.039                                       \\
\multicolumn{1}{l|}{AFNet~\cite{feng2019attentive}}             & \multicolumn{1}{c|}{2019}                   & \multicolumn{1}{c|}{VGG16}                      & \multicolumn{1}{c|}{{\color[HTML]{3166FF} \textbf{21.08}}}      & 0.793                                       & 0.785                                       & 0.895                                       & \multicolumn{1}{c|}{0.046}                                       & 0.908                                       & 0.886                                       & {\color[HTML]{3166FF} \textbf{0.941}}       & \multicolumn{1}{c|}{{\color[HTML]{3166FF} \textbf{0.042}}}       & 0.738                                       & {\color[HTML]{3166FF} \textbf{0.717}}       & 0.859                                       & \multicolumn{1}{c|}{{\color[HTML]{FE0000} \textbf{0.057}}}       & 0.828                                       & {\color[HTML]{3166FF} \textbf{0.804}}       & {\color[HTML]{FE0000} \textbf{0.887}}       & \multicolumn{1}{c|}{{\color[HTML]{FE0000} \textbf{0.071}}}       & 0.888                                       & 0.869                                       & {\color[HTML]{3166FF} \textbf{0.947}}       & {\color[HTML]{3166FF} \textbf{0.036}}       \\
\multicolumn{1}{l|}{CPD~\cite{wu2019cascaded}}               & \multicolumn{1}{c|}{2019}                   & \multicolumn{1}{c|}{VGG16}                      & \multicolumn{1}{c|}{29.32}                                      & {\color[HTML]{3166FF} \textbf{0.813}}       & {\color[HTML]{FE0000} \textbf{0.801}}       & {\color[HTML]{FE0000} \textbf{0.908}}       & \multicolumn{1}{c|}{{\color[HTML]{FE0000} \textbf{0.043}}}       & {\color[HTML]{3166FF} \textbf{0.914}}       & {\color[HTML]{FE0000} \textbf{0.895}}       & {\color[HTML]{FE0000} \textbf{0.943}}       & \multicolumn{1}{c|}{{\color[HTML]{FE0000} \textbf{0.040}}}       & {\color[HTML]{FE0000} \textbf{0.745}}       & 0.715                                       & {\color[HTML]{FE0000} \textbf{0.869}}       & \multicolumn{1}{c|}{{\color[HTML]{FE0000} \textbf{0.057}}}       & {\color[HTML]{FE0000} \textbf{0.832}}       & {\color[HTML]{FE0000} \textbf{0.806}}       & {\color[HTML]{3166FF} \textbf{0.884}}       & \multicolumn{1}{c|}{{\color[HTML]{3166FF} \textbf{0.074}}}       & {\color[HTML]{3166FF} \textbf{0.895}}       & {\color[HTML]{3166FF} \textbf{0.879}}       & {\color[HTML]{FE0000} \textbf{0.950}}       & {\color[HTML]{FE0000} \textbf{0.033}}       \\
\multicolumn{1}{l|}{\textbf{CAGNet-V}} & \multicolumn{1}{c|}{-}                      & \multicolumn{1}{c|}{VGG16}                      & \multicolumn{1}{c|}{{\color[HTML]{FE0000} \textbf{20.98}}}      & {\color[HTML]{FE0000} \textbf{0.823}}       & {\color[HTML]{3166FF} \textbf{0.797}}       & {\color[HTML]{3166FF} \textbf{0.904}}       & \multicolumn{1}{c|}{{\color[HTML]{3166FF} \textbf{0.044}}}       & {\color[HTML]{FE0000} \textbf{0.915}}       & {\color[HTML]{3166FF} \textbf{0.893}}       & 0.939                                       & \multicolumn{1}{c|}{{\color[HTML]{3166FF} \textbf{0.042}}}       & {\color[HTML]{3166FF} \textbf{0.744}}       & {\color[HTML]{FE0000} \textbf{0.718}}       & {\color[HTML]{3166FF} \textbf{0.860}}       & \multicolumn{1}{c|}{{\color[HTML]{FE0000} \textbf{0.057}}}       & {\color[HTML]{3166FF} \textbf{0.831}}       & 0.799                                       & 0.881                                       & \multicolumn{1}{c|}{0.077}                                       & {\color[HTML]{FE0000} \textbf{0.906}}       & {\color[HTML]{FE0000} \textbf{0.886}}       & {\color[HTML]{3166FF} \textbf{0.947}}       & {\color[HTML]{FE0000} \textbf{0.033}}       \\ \hline
\multicolumn{24}{c}{ResNet~\cite{he2016deep}}                                                                                                                                                                                                                                                                                                                                                                                                                                                                                                                                                                                                                                                                                                                                                                                                                                                                                                                                                                                                                                                                                                                                                                                                                   \\ \hline
\multicolumn{1}{l|}{SRM~\cite{wang2017stagewise}}               & \multicolumn{1}{c|}{2017}                   & \multicolumn{1}{c|}{ResNet50}                   & \multicolumn{1}{c|}{43.74}                                      & 0.753                                       & 0.722                                       & 0.867                                       & \multicolumn{1}{c|}{0.059}                                       & 0.892                                       & 0.853                                       & 0.927                                       & \multicolumn{1}{c|}{0.054}                                       & 0.707                                       & 0.658                                       & 0.843                                       & \multicolumn{1}{c|}{0.069}                                       & 0.803                                       & 0.762                                       & 0.861                                       & \multicolumn{1}{c|}{0.087}                                       & 0.874                                       & 0.835                                       & 0.938                                       & 0.046                                       \\
\multicolumn{1}{l|}{DGRL~\cite{wang2018detect}}              & \multicolumn{1}{c|}{2018}                   & \multicolumn{1}{c|}{ResNet50}                   & \multicolumn{1}{c|}{126.35}                                     & 0.794                                       & 0.774                                       & 0.899                                       & \multicolumn{1}{c|}{0.050}                                       & 0.906                                       & 0.891                                       & {\color[HTML]{3166FF} \textbf{0.946}}       & \multicolumn{1}{c|}{0.041}                                       & 0.733                                       & 0.709                                       & 0.856                                       & \multicolumn{1}{c|}{0.062}                                       & 0.827                                       & 0.802                                       & {\color[HTML]{3166FF} \textbf{0.891}}       & \multicolumn{1}{c|}{0.073}                                       & 0.890                                       & 0.875                                       & 0.949                                       & 0.036                                       \\
\multicolumn{1}{l|}{PiCANet-R~\cite{liu2018picanet}}         & \multicolumn{1}{c|}{2018}                   & \multicolumn{1}{c|}{ResNet50}                   & \multicolumn{1}{c|}{{\color[HTML]{3166FF} \textbf{37.02}}}      & 0.759                                       & 0.755                                       & 0.873                                       & \multicolumn{1}{c|}{0.051}                                       & 0.886                                       & 0.867                                       & 0.927                                       & \multicolumn{1}{c|}{0.046}                                       & 0.717                                       & 0.695                                       & 0.848                                       & \multicolumn{1}{c|}{0.065}                                       & 0.804                                       & 0.782                                       & 0.862                                       & \multicolumn{1}{c|}{0.078}                                       & 0.870                                       & 0.840                                       & 0.940                                       & 0.043                                       \\
\multicolumn{1}{l|}{CapSal~\cite{zhang2019capsal}}            & \multicolumn{1}{c|}{2019}                   & \multicolumn{1}{c|}{ResNet101}                  & \multicolumn{1}{c|}{91.09}                                      & 0.755                                       & 0.689                                       & 0.867                                       & \multicolumn{1}{c|}{0.063}                                       & ---                                         & ---                                         & ---                                         & \multicolumn{1}{c|}{---}                                         & ---                                         & ---                                         & ---                                         & \multicolumn{1}{c|}{---}                                         & 0.827                                       & 0.791                                       & 0.878                                       & \multicolumn{1}{c|}{0.074}                                       & 0.841                                       & 0.780                                       & 0.905                                       & 0.058                                       \\
\multicolumn{1}{l|}{BASNet~\cite{qin2019basnet}}            & \multicolumn{1}{c|}{2019}                   & \multicolumn{1}{c|}{ResNet34}                   & \multicolumn{1}{c|}{87.06}                                      & 0.791                                       & {\color[HTML]{3166FF} \textbf{0.803}}       & 0.884                                       & \multicolumn{1}{c|}{0.047}                                       & 0.880                                       & {\color[HTML]{FE0000} \textbf{0.904}}       & 0.921                                       & \multicolumn{1}{c|}{{\color[HTML]{FE0000} \textbf{0.037}}}       & {\color[HTML]{FE0000} \textbf{0.756}}       & {\color[HTML]{FE0000} \textbf{0.751}}       & {\color[HTML]{3166FF} \textbf{0.869}}       & \multicolumn{1}{c|}{{\color[HTML]{3166FF} \textbf{0.056}}}       & 0.781                                       & 0.800                                       & 0.853                                       & \multicolumn{1}{c|}{0.077}                                       & {\color[HTML]{3166FF} \textbf{0.895}}       & {\color[HTML]{3166FF} \textbf{0.889}}       & 0.946                                       & {\color[HTML]{3166FF} \textbf{0.032}}       \\
\multicolumn{1}{l|}{CPD-R~\cite{wu2019cascaded}}             & \multicolumn{1}{c|}{2019}                   & \multicolumn{1}{c|}{ResNet50}                   & \multicolumn{1}{c|}{47.85}                                      & {\color[HTML]{3166FF} \textbf{0.805}}       & 0.795                                       & {\color[HTML]{3166FF} \textbf{0.904}}       & \multicolumn{1}{c|}{{\color[HTML]{3166FF} \textbf{0.043}}}       & {\color[HTML]{3166FF} \textbf{0.917}}       & 0.898                                       & {\color[HTML]{FE0000} \textbf{0.949}}       & \multicolumn{1}{c|}{{\color[HTML]{FE0000} \textbf{0.037}}}       & 0.747                                       & 0.719                                       & {\color[HTML]{FE0000} \textbf{0.873}}       & \multicolumn{1}{c|}{{\color[HTML]{3166FF} \textbf{0.056}}}       & {\color[HTML]{3166FF} \textbf{0.831}}       & {\color[HTML]{3166FF} \textbf{0.803}}       & 0.887                                       & \multicolumn{1}{c|}{{\color[HTML]{3166FF} \textbf{0.072}}}       & 0.891                                       & 0.875                                       & {\color[HTML]{FE0000} \textbf{0.950}}       & 0.034                                       \\
\multicolumn{1}{l|}{\textbf{CAGNet-R}} & \multicolumn{1}{c|}{-}                      & \multicolumn{1}{c|}{ResNet50}                   & \multicolumn{1}{c|}{{\color[HTML]{FE0000} \textbf{26.06}}}      & {\color[HTML]{FE0000} \textbf{0.838}}       & {\color[HTML]{FE0000} \textbf{0.817}}       & {\color[HTML]{FE0000} \textbf{0.914}}       & \multicolumn{1}{c|}{{\color[HTML]{FE0000} \textbf{0.040}}}       & {\color[HTML]{FE0000} \textbf{0.921}}       & {\color[HTML]{3166FF} \textbf{0.903}}       & 0.944                                       & \multicolumn{1}{c|}{{\color[HTML]{FE0000} \textbf{0.037}}}       & {\color[HTML]{3166FF} \textbf{0.753}}       & {\color[HTML]{3166FF} \textbf{0.729}}       & 0.862                                       & \multicolumn{1}{c|}{{\color[HTML]{FE0000} \textbf{0.054}}}       & {\color[HTML]{FE0000} \textbf{0.847}}       & {\color[HTML]{FE0000} \textbf{0.820}}       & {\color[HTML]{FE0000} \textbf{0.896}}       & \multicolumn{1}{c|}{{\color[HTML]{FE0000} \textbf{0.067}}}       & {\color[HTML]{FE0000} \textbf{0.910}}       & {\color[HTML]{FE0000} \textbf{0.893}}       & {\color[HTML]{FE0000} \textbf{0.950}}       & {\color[HTML]{FE0000} \textbf{0.030}}       \\ \hline
\multicolumn{24}{c}{NASNet~\cite{zoph2018learning}}                                                                                                                                                                                                                                                                                                                                                                                                                                                                                                                                                                                                                                                                                                                                                                                                                                                                                                                                                                                                                                                                                                                                                                                                                    \\ \hline
\multicolumn{1}{l|}{\textbf{CAGNet-M}} & \multicolumn{1}{c|}{-}                      & \multicolumn{1}{c|}{Mobile}                     & \multicolumn{1}{c|}{{\color[HTML]{FE0000} {\ul \textbf{5.57}}}} & {\color[HTML]{3166FF} \textbf{0.852}}       & {\color[HTML]{3166FF} \textbf{0.832}}       & {\color[HTML]{3166FF} \textbf{0.923}}       & \multicolumn{1}{c|}{{\color[HTML]{3166FF} \textbf{0.037}}}       & {\color[HTML]{3166FF} \textbf{0.933}}       & {\color[HTML]{3166FF} \textbf{0.916}}       & {\color[HTML]{3166FF} \textbf{0.953}}       & \multicolumn{1}{c|}{{\color[HTML]{3166FF} \textbf{0.034}}}       & {\color[HTML]{3166FF} \textbf{0.764}}       & {\color[HTML]{3166FF} \textbf{0.743}}       & {\color[HTML]{3166FF} \textbf{0.864}}       & \multicolumn{1}{c|}{{\color[HTML]{3166FF} \textbf{0.052}}}       & {\color[HTML]{3166FF} \textbf{0.846}}       & {\color[HTML]{3166FF} \textbf{0.819}}       & {\color[HTML]{3166FF} \textbf{0.893}}       & \multicolumn{1}{c|}{{\color[HTML]{3166FF} \textbf{0.069}}}       & {\color[HTML]{3166FF} \textbf{0.919}}       & {\color[HTML]{3166FF} \textbf{0.904}}       & {\color[HTML]{3166FF} \textbf{0.956}}       & {\color[HTML]{3166FF} \textbf{0.028}}       \\
\multicolumn{1}{l|}{\textbf{CAGNet-L}} & \multicolumn{1}{c|}{-}                      & \multicolumn{1}{c|}{Large}                      & \multicolumn{1}{c|}{89.42}                                      & {\color[HTML]{FE0000} {\ul \textbf{0.886}}} & {\color[HTML]{FE0000} {\ul \textbf{0.871}}} & {\color[HTML]{FE0000} {\ul \textbf{0.944}}} & \multicolumn{1}{c|}{{\color[HTML]{FE0000} {\ul \textbf{0.029}}}} & {\color[HTML]{FE0000} {\ul \textbf{0.943}}} & {\color[HTML]{FE0000} {\ul \textbf{0.932}}} & {\color[HTML]{FE0000} {\ul \textbf{0.963}}} & \multicolumn{1}{c|}{{\color[HTML]{FE0000} {\ul \textbf{0.026}}}} & {\color[HTML]{FE0000} {\ul \textbf{0.798}}} & {\color[HTML]{FE0000} {\ul \textbf{0.779}}} & {\color[HTML]{FE0000} {\ul \textbf{0.889}}} & \multicolumn{1}{c|}{{\color[HTML]{FE0000} {\ul \textbf{0.047}}}} & {\color[HTML]{FE0000} {\ul \textbf{0.877}}} & {\color[HTML]{FE0000} {\ul \textbf{0.858}}} & {\color[HTML]{FE0000} {\ul \textbf{0.921}}} & \multicolumn{1}{c|}{{\color[HTML]{FE0000} {\ul \textbf{0.053}}}} & {\color[HTML]{FE0000} {\ul \textbf{0.932}}} & {\color[HTML]{FE0000} {\ul \textbf{0.921}}} & {\color[HTML]{FE0000} {\ul \textbf{0.965}}} & {\color[HTML]{FE0000} {\ul \textbf{0.024}}} \\ \hline
\end{tabular}
}
}
\end{table}

\begin{table}[!t]
\centering
\caption{Attributes-based performance on the challenging SOC dataset. The best score and the second best score are shown in {\color[HTML]{FE0000}\textbf{red}} and {\color[HTML]{3166FF}\textbf{blue}}, respectively,}
\label{tab:attrs}
\resizebox{\textwidth}{!}{%
{\renewcommand{\arraystretch}{1.6}
\begin{tabular}{l|cc|cc|cc|cc|cc|cc|cc|cc|cc|cc|}
\hline
Attribute         & \multicolumn{2}{c|}{AC}                                                       & \multicolumn{2}{c|}{BO}                                                       & \multicolumn{2}{c|}{CL}                                                       & \multicolumn{2}{c|}{HO}                                                       & \multicolumn{2}{c|}{MB}                                                       & \multicolumn{2}{c|}{OC}                                                       & \multicolumn{2}{c|}{OV}                                                       & \multicolumn{2}{c|}{SC}                                                       & \multicolumn{2}{c|}{SO}                                                       & \multicolumn{2}{c|}{average}                                                  \\ \hline
Metric            & avgF                                  & E                                     & avgF                                  & E                                     & avgF                                  & E                                     & avgF                                  & E                                     & avgF                                  & E                                     & avgF                                  & E                                     & avgF                                  & E                                     & avgF                                  & E                                     & avgF                                  & E                                     & avgF                                  & E                                     \\ \hline
Amulet~\cite{zhang2017amulet}            & 0.682                                 & 0.782                                 & 0.749                                 & 0.509                                 & 0.668                                 & 0.744                                 & 0.689                                 & 0.774                                 & 0.700                                 & 0.809                                 & 0.668                                 & 0.738                                 & 0.746                                 & 0.772                                 & 0.640                                 & 0.749                                 & 0.513                                 & 0.675                                 & 0.673                                 & 0.728                                 \\
NLDF~\cite{luo2017non}              & 0.697                                 & 0.812                                 & 0.638                                 & 0.448                                 & 0.679                                 & 0.738                                 & 0.706                                 & 0.795                                 & 0.693                                 & 0.807                                 & 0.658                                 & 0.749                                 & 0.718                                 & 0.755                                 & 0.670                                 & 0.790                                 & 0.554                                 & 0.725                                 & 0.665                                 & 0.736                                 \\
DSS~\cite{hou2017deeply}               & 0.663                                 & 0.778                                 & 0.682                                 & 0.450                                 & 0.633                                 & 0.718                                 & 0.677                                 & 0.772                                 & 0.706                                 & 0.806                                 & 0.630                                 & 0.725                                 & 0.689                                 & 0.727                                 & 0.641                                 & 0.766                                 & 0.540                                 & 0.720                                 & 0.651                                 & 0.718                                 \\
SRM~\cite{wang2017stagewise}               & 0.734                                 & 0.834                                 & 0.785                                 & 0.567                                 & 0.714                                 & 0.781                                 & 0.737                                 & 0.818                                 & 0.793                                 & 0.867                                 & 0.687                                 & 0.772                                 & 0.781                                 & 0.800                                 & 0.681                                 & 0.801                                 & 0.585                                 & 0.752                                 & 0.722                                 & 0.777                                 \\
BMPM~\cite{zhang2018bi}              & 0.712                                 & 0.815                                 & 0.530                                 & 0.392                                 & 0.657                                 & 0.745                                 & 0.704                                 & 0.802                                 & 0.734                                 & 0.846                                 & 0.670                                 & 0.769                                 & 0.729                                 & 0.771                                 & 0.670                                 & 0.803                                 & 0.568                                 & 0.757                                 & 0.664                                 & 0.745                                 \\
BASNet~\cite{qin2019basnet}            & 0.740                                 & 0.844                                 & 0.503                                 & 0.433                                 & 0.679                                 & 0.766                                 & 0.702                                 & 0.796                                 & 0.805                                 & 0.868                                 & 0.654                                 & 0.760                                 & 0.719                                 & 0.764                                 & 0.655                                 & 0.802                                 & 0.590                                 & 0.766                                 & 0.672                                 & 0.755                                 \\
CPD-R~\cite{wu2019cascaded}             & 0.765                                 & {\color[HTML]{3166FF} \textbf{0.860}} & {\color[HTML]{3166FF} \textbf{0.824}} & 0.662                                 & 0.741                                 & 0.808                                 & 0.766                                 & 0.845                                 & 0.812                                 & 0.880                                 & 0.741                                 & 0.813                                 & 0.799                                 & {\color[HTML]{3166FF} \textbf{0.831}} & 0.727                                 & 0.837                                 & 0.635                                 & 0.796                                 & 0.757                                 & 0.815                                 \\
\textbf{CAGNet-V} & {\color[HTML]{3166FF} \textbf{0.769}} & 0.852                                 & 0.772                                 & 0.611                                 & 0.732                                 & 0.795                                 & 0.775                                 & 0.851                                 & 0.813                                 & 0.886                                 & 0.729                                 & 0.809                                 & 0.791                                 & 0.817                                 & 0.739                                 & 0.841                                 & 0.630                                 & 0.794                                 & 0.750                                 & 0.806                                 \\
\textbf{CAGNet-R} & 0.741                                 & 0.848                                 & 0.799                                 & {\color[HTML]{3166FF} \textbf{0.675}} & 0.726                                 & 0.791                                 & 0.764                                 & 0.835                                 & {\color[HTML]{3166FF} \textbf{0.849}} & {\color[HTML]{3166FF} \textbf{0.899}} & 0.727                                 & 0.803                                 & 0.781                                 & 0.807                                 & 0.747                                 & {\color[HTML]{3166FF} \textbf{0.856}} & 0.633                                 & 0.790                                 & 0.752                                 & 0.812                                 \\
\textbf{CAGNet-M} & 0.765                                 & 0.856                                 & 0.784                                 & 0.649                                 & {\color[HTML]{3166FF} \textbf{0.743}} & {\color[HTML]{3166FF} \textbf{0.814}} & {\color[HTML]{3166FF} \textbf{0.795}} & {\color[HTML]{3166FF} \textbf{0.861}} & 0.825                                 & {\color[HTML]{3166FF} \textbf{0.899}} & {\color[HTML]{3166FF} \textbf{0.747}} & {\color[HTML]{FE0000} \textbf{0.833}} & {\color[HTML]{3166FF} \textbf{0.801}} & 0.828                                 & {\color[HTML]{3166FF} \textbf{0.752}} & 0.851                                 & {\color[HTML]{3166FF} \textbf{0.647}} & {\color[HTML]{3166FF} \textbf{0.800}} & {\color[HTML]{3166FF} \textbf{0.762}} & {\color[HTML]{3166FF} \textbf{0.821}} \\
\textbf{CAGNet-L} & {\color[HTML]{FE0000} \textbf{0.785}} & {\color[HTML]{FE0000} \textbf{0.873}} & {\color[HTML]{FE0000} \textbf{0.827}} & {\color[HTML]{FE0000} \textbf{0.719}} & {\color[HTML]{FE0000} \textbf{0.754}} & {\color[HTML]{FE0000} \textbf{0.827}} & {\color[HTML]{FE0000} \textbf{0.816}} & {\color[HTML]{FE0000} \textbf{0.876}} & {\color[HTML]{FE0000} \textbf{0.879}} & {\color[HTML]{FE0000} \textbf{0.925}} & {\color[HTML]{FE0000} \textbf{0.749}} & {\color[HTML]{3166FF} \textbf{0.829}} & {\color[HTML]{FE0000} \textbf{0.828}} & {\color[HTML]{FE0000} \textbf{0.852}} & {\color[HTML]{FE0000} \textbf{0.780}} & {\color[HTML]{FE0000} \textbf{0.881}} & {\color[HTML]{FE0000} \textbf{0.692}} & {\color[HTML]{FE0000} \textbf{0.829}} & {\color[HTML]{FE0000} \textbf{0.790}} & {\color[HTML]{FE0000} \textbf{0.846}} \\ \hline
\end{tabular}
}
}
\end{table}

{\it Quantitative Evaluation.} P-R curves and F-measure curves on the five datasets are shown in Figure~\ref{fig:PRcurves} and Figure~\ref{fig:Fcurves}, respectively. We can see that our proposed method performs favorably against other methods in all cases. Especially, it is obvious that our  CAGNet-L performs better than all other methods by a relatively large margin. Moreover, we compare our method with other previous state-of-the-art methods in terms of avgF, wF, E-measure, and MAE score on five benchmark datasets in Table~\ref{tab:comparison}. As seen from this table, our method ranks first in most cases. We also evaluate attributes-based performance on the challenging SOC dataset. The results of the recent state-of-the-art methods on nine attributes of SOC and their average are shown in Table~\ref{tab:attrs}. We can see that our method has achieved great performance and is capable of handling the complex scenarios of SOC dataset. It is interesting to note that our method contains fewer parameters than the existing ones, is end-to-end, and does not need any post-processing steps such as CRF~\cite{krahenbuhl2011efficient}. Another interesting thing about our method is that although our CAGNet-M has significantly fewer parameters than the other networks (only 5.57 million parameters), it has shown outstanding performance. This functionality is desirable for the applications in which we have limitation in terms of the memory. Furthermore, our CAGNet-V has a real-time speed of $28$ FPS when processing a $480\times 480$ image, and therefore it can be practically adopted as a preprocessing step for computer vision tasks.

\begin{figure}[!t]
\begin{center}
\includegraphics[width=1\textwidth, height =6.7cm]{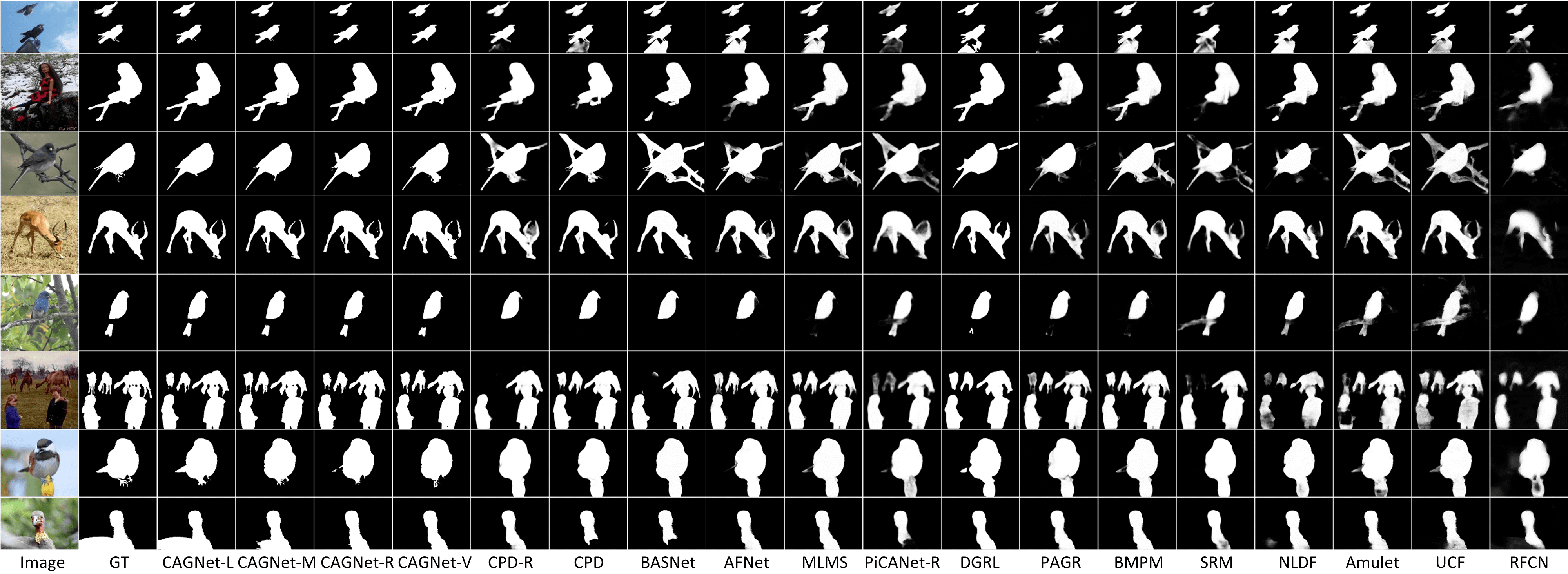}
\end{center}
  \caption{Qualitative comparisons with previous state-of-the-art methods. As it can be seen, our method is capable of predicting saliency maps that are closer to the ground truth compared to the other methods.}
\label{fig:SM_VC}
\end{figure}

{\it Qualitative Evaluation.} Some qualitative results are shown in Figure~\ref{fig:SM_VC}. Thanks to the proposed modules, it can be seen that our model is capable of highlighting the inner part of foreground regions in various complicated scenes. Furthermore, our model is able to suppress the background regions which are incorrectly labeled by other saliency detection methods. Thus, by taking advantage of different proposed modules, our method is able to handle various complex scenarios.

\subsection{Ablation analysis}
Our proposed CAGNet consists of three modules, namely the Multi-scale Feature Extraction Module (MFEM), the Guide Module, and the Residual Refinement Module (RRM). We perform the ablation analysis on CAGNet-V by using three challenging large-scale datasets, namely DUTS-TE~\cite{wang2017learning}, DUT-O~\cite{yang2013saliency}, and HKU-IS~\cite{li2015visual}.  In order to investigate the effectiveness of each module, we gradually add them to our base network. Our base network is obtained by applying the following modifications to the CAGNet: i) replacing the MFEM modules with $1 \times 1$ convolutions with the same number of filters, ii) removing Guide Modules from the network (which means that the multi-level features are not multiplied by the channel weights and the spatial weights), iii) removing the RRM modules from the model. 

We perform ablation analysis by adding each module to our base network in a stepwise manner. The results are shown in Table~\ref{tab:abl}. In this table, the base network is denoted as Base.

\begin{table}[!t]
\centering
\caption{Ablation analysis of our proposed method with different settings. The best results are shown in {\color[HTML]{FE0000} \textbf{red}}.}
\label{tab:abl}
\resizebox{0.85\linewidth}{!}{%
{\renewcommand{\arraystretch}{1.4}
\begin{tabular}{l|ccc|ccc|ccc}
\hline
Dataset                                                                                & \multicolumn{3}{c|}{DUTS-TE~\cite{wang2017learning}}                                                                                             & \multicolumn{3}{c|}{DUT-O~\cite{yang2013saliency}}                                                                                               & \multicolumn{3}{c}{HKU-IS~\cite{li2015visual}}                                                                                              \\ \hline
Metric                                                                                 & avgF                                   & wF                                     & MAE                                    & avgF                                   & wF                                     & MAE                                    & avgF                                   & wF                                     & MAE                                    \\ \hline
Base                                                                                   & 0.7452                                 & 0.7099                                 & 0.0627                                 & 0.6481                                 & 0.6027                                 & 0.0862                                 & 0.8579                                 & 0.8298                                 & 0.0482                                 \\
Base + HG                                                                              & 0.7598                                 & 0.7204                                 & 0.0588                                 & 0.6571                                 & 0.6068                                 & 0.0819                                 & 0.8700                                 & 0.8398                                 & 0.0450                                 \\
Base + LG                                                                              & 0.7650                                 & 0.7268                                 & 0.0588                                 & 0.6649                                 & 0.6178                                 & 0.0817                                 & 0.8714                                 & 0.8419                                 & 0.0447                                 \\
Base + GM                                                                              & 0.7707                                 & 0.7335                                 & 0.0558                                 & 0.6687                                 & 0.6209                                 & 0.0780                                 & 0.8743                                 & 0.8451                                 & 0.0432                                 \\
Base + GM + MFEM                                                                       & 0.8003                                 & 0.7779                                 & 0.0481                                 & 0.7256                                 & 0.6971                                 & 0.0616                                 & 0.8960                                 & 0.8776                                 & 0.0346                                 \\
\textbf{\begin{tabular}[c]{@{}l@{}}Base + GM + MFEM + RRM\\ (= CAGNet-V)\end{tabular}} & {\color[HTML]{FE0000} \textbf{0.8226}} & {\color[HTML]{FE0000} \textbf{0.7971}} & {\color[HTML]{FE0000} \textbf{0.0445}} & {\color[HTML]{FE0000} \textbf{0.7444}} & 0.7179                                 & {\color[HTML]{FE0000} \textbf{0.0571}} & {\color[HTML]{FE0000} \textbf{0.9056}} & {\color[HTML]{FE0000} \textbf{0.8858}} & 0.0332                                 \\ \hline
CE Loss Function                                                                       & 0.7591                                 & 0.7517                                 & 0.0524                                 & 0.7017                                 & 0.6793                                 & 0.0652                                 & 0.8783                                 & 0.8558                                 & 0.0398                                 \\
Dilated Convolution                                                                    & 0.8214                                 & 0.7961                                 & 0.0457                                 & 0.7414                                 & 0.7152                                 & 0.0583                                 & 0.9029                                 & 0.8811                                 & 0.0343                                 \\
Trivial Convolution                                                                    & 0.8166                                 & 0.7940                                 & 0.0458                                 & 0.7439                                 & {\color[HTML]{FE0000} \textbf{0.7203}} & 0.0581                                 & 0.9041                                 & 0.8857                                 & {\color[HTML]{FE0000} \textbf{0.0331}} \\ \hline
\end{tabular}
}
}
\end{table}

\begin{figure}[!t]
\begin{center}
\includegraphics[width=0.5\linewidth]{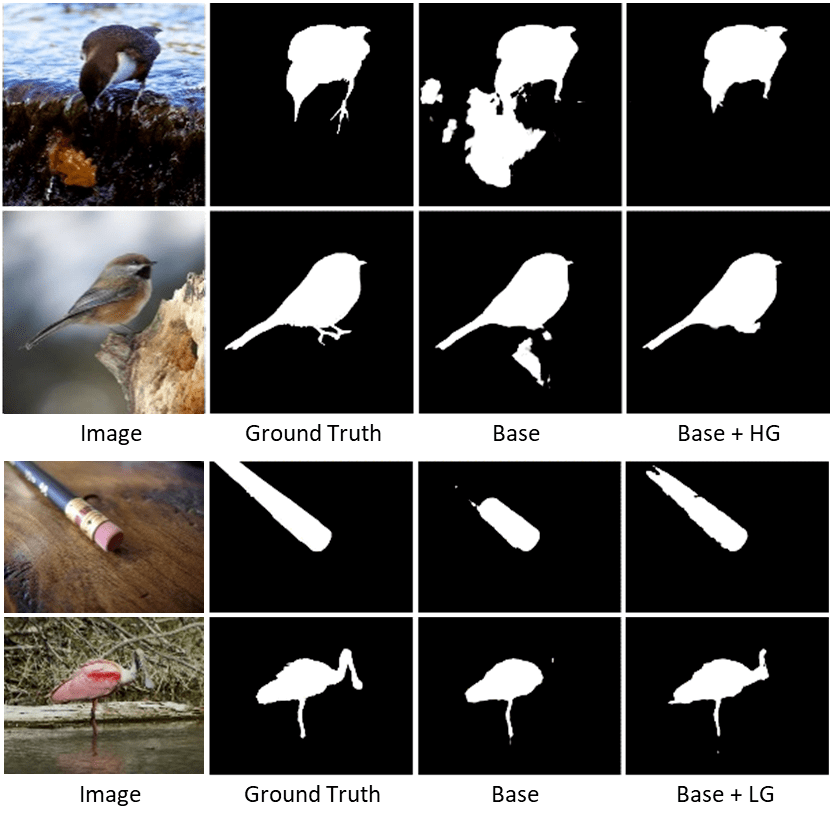}
\end{center}
\caption{Visual Comparison for two branches of our Guide Module. The first and second rows show the comparison for High-level Guide branch (denoted as HG). The third and fourth rows show the comparison for Low-level Guide branch (denoted as LG).}
\label{fig:SM_abl}
\end{figure}

{\it The effectiveness of Guide Module.} We add the High-level Guide branch, Low-level Guide branch, and both High-level and Low-level Guide branches (i.e., the Guide Module) to the base network, which are denoted as HG, LG, and GM, respectively in Table~\ref{tab:abl}. As seen from this table, the performance improves, which shows the beneficial effect of using our Guide Module. Using this module results in i) making salient and non-salient regions more distinct and suppressing the non-salient regions that have "salient-like" appearance, ii) assigning foreground label to different-looking salient regions. To further investigate the effectiveness of our guide branches, we show a visual comparison for each branch in Figure~\ref{fig:SM_abl}. As it can be seen, when we add the High-level Guide branch to the base network (Base+HG), the non-salient regions that have "salient-like" appearance are suppressed. Furthermore, when we add the Low-level Guide branch to the base network (Base+LG), different-looking salient regions (head of the pencil and the rest of the pencil, head of the bird and the rest of the bird) are labeled as salient.

{\it The effectiveness of MFEM.} Based on the aforementioned architecture, we replace the $1 \times 1$ convolutions with the MFEM modules. As seen in Table~\ref{tab:abl}, our proposed MFEM has a beneficial effect on saliency detection and improves the results, which shows extracting multi-scale features can help to detect salient objects with different scales and locations. 

{\it The effectiveness of RRM.} To reveal the effect of the RRMs, we add them to the aforementioned architecture. From Table~\ref{tab:abl}, it can be observed that using our refinement module is helpful for saliency detection and improves the performance. 

{\it The effectiveness of our designed loss.} To demonstrate the effectiveness of our designed loss function, we train our CAGNet-V with Cross-entropy, which is denoted as CE Loss Function in Table~\ref{tab:abl}. As seen in this table, our designed loss outperforms the cross-entropy loss by a significant margin.

\begin{table}[!t]
\centering
\caption{The results of CAGNet-V with different settings for the parameter N. The best results are shown in {\color[HTML]{FE0000} \textbf{red}}. The unit of the total number of parameters (denoted as \#Par) is million.}
\label{tab:channels}
\resizebox{0.8\linewidth}{!}{%
{\renewcommand{\arraystretch}{1.6}
\begin{tabular}{l|ccc|ccc|ccc|c}
\hline
Dataset      & \multicolumn{3}{c|}{DUTS-TE~\cite{wang2017learning}}                                                                                             & \multicolumn{3}{c|}{DUT-O~\cite{yang2013saliency}}                                                                                               & \multicolumn{3}{c|}{HKU-IS~\cite{li2015visual}}                                                                                              &                       \\ \cline{1-10}
Metric       & avgF                                   & wF                                     & MAE                                    & avgF                                   & wF                                     & MAE                                    & avgF                                   & wF                                     & MAE                                    & \multirow{-2}{*}{\#Par} \\ \hline
N=1          & 0.8068                                 & 0.7777                                 & 0.0474                                 & 0.7254                                 & 0.6932                                 & 0.0611                                 & 0.8963                                 & 0.8734                                 & 0.0355                                 & 19.61                 \\
N=2          & 0.8100                                 & 0.7821                                 & 0.0469                                 & 0.7339                                 & 0.7044                                 & 0.0601                                 & 0.9020                                 & 0.8811                                 & 0.0337                                 & 19.79                 \\
N=4          & 0.8155                                 & 0.7903                                 & 0.0461                                 & 0.7412                                 & 0.7145                                 & 0.0576                                 & 0.9010                                 & 0.8807                                 & 0.0336                                 & 20.17                 \\
\textbf{N=8} & {\color[HTML]{FE0000} \textbf{0.8226}} & 0.7971                                 & {\color[HTML]{FE0000} \textbf{0.0445}} & 0.7444                                 & 0.7179                                 & 0.0571                                 & {\color[HTML]{FE0000} \textbf{0.9056}} & {\color[HTML]{FE0000} \textbf{0.8858}} & 0.0332                                 & 20.98                 \\
N=16         & 0.8189                                 & 0.7935                                 & 0.0451                                 & 0.7474                                 & 0.7205                                 & 0.0567                                 & 0.9050                                 & 0.8848                                 & {\color[HTML]{FE0000} \textbf{0.0328}} & 22.86                 \\
N=32         & 0.8218                                 & {\color[HTML]{FE0000} \textbf{0.7973}} & 0.0452                                 & 0.7497                                 & 0.7255                                 & 0.0576                                 & 0.9048                                 & 0.8843                                 & 0.0332                                 & 27.61                 \\
N=64         & 0.8189                                 & 0.7932                                 & 0.0469                                 & {\color[HTML]{FE0000} \textbf{0.7523}} & {\color[HTML]{FE0000} \textbf{0.7278}} & {\color[HTML]{FE0000} \textbf{0.0561}} & 0.8980                                 & 0.8773                                 & 0.0354                                 & 41.04                 \\ \hline
\end{tabular}%
}
}
\end{table}

To further prove the effectiveness of our MFEM, we implement the MFEMs in CAGNet-V by adopting dilated convolutional layers (kernel size=3, dilation rates=1, 3, 5, 7), denoted as Dilated Convolution in Table~\ref{tab:abl}. We can see that the performance degrades, which shows that our proposed MFEM can capture more powerful multi-scale features by enabling densely connections within a large $k \times k$ region in the feature map. We also implement the MFEMs in CAGNet-V by adopting trivial convolutional layers with kernel size=3, 7, 11, 15, denoted as Trivial Convolution in Table~\ref{tab:abl}. As seen from this table, the performance gets worse compared to our CAGNet-V with the proposed MFEM. It is interesting to note that CAGNet-V with our proposed MFEM contains fewer parameters than the CAGNet-V with the MFEM implemented by adopting trivial convolutional layers (20.98 million vs. 27.03 million), which is due to the architectural design of GCNs.

We perform another experiment on CAGNet-V and train it with different settings for the parameter $N$. The results are shown in Table~\ref{tab:channels}. In this paper, by considering the trade-off between the performance and the number of parameters, we have chosen $N=8$ for our method.

\begin{figure}[!t]
\begin{center}
\includegraphics[width=1\linewidth]{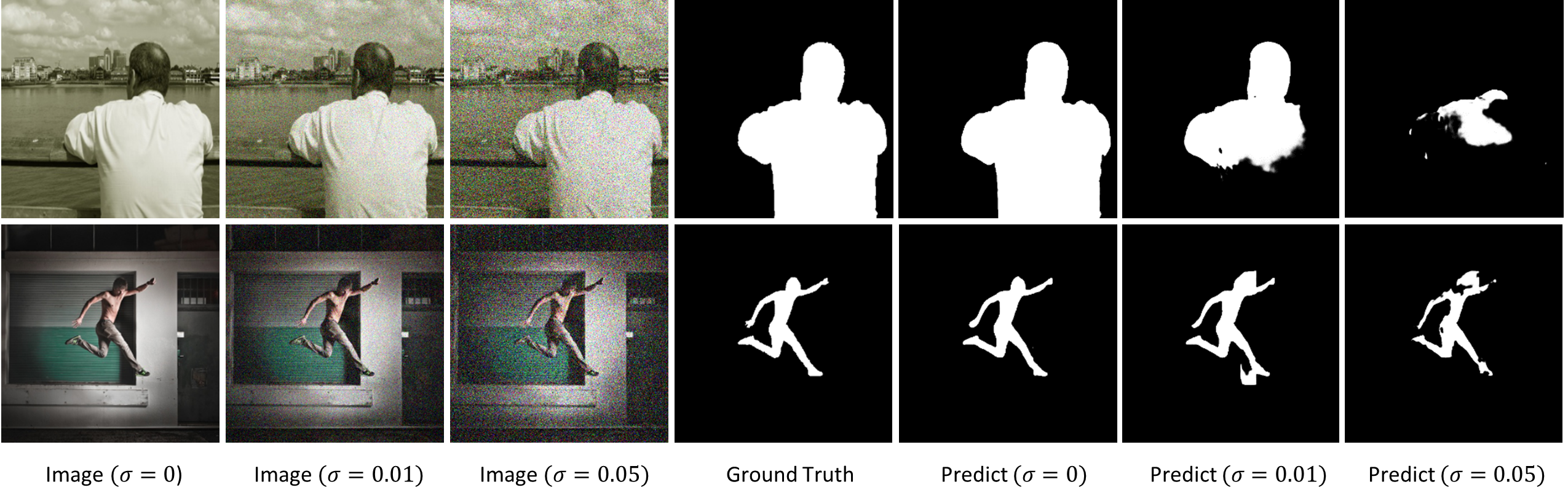}
\end{center}
\caption{The predictions of our method for two images corrupted by Additive White Gaussian Noise (AWGN) with two variance values ($\sigma=0.01$ and $\sigma=0.05$).}
\label{fig:SM_Noise}
\end{figure}

\section{Conclusion and future work}
In this paper, we propose a novel end-to-end framework that has the power of i) making the foreground and background regions more distinct and suppressing the non-salient regions that have "salient-like" appearance; ii) detecting salient objects that have different-looking regions. Our proposed model is also capable of capturing multi-scale contextual information effectively. The attentive guided multi-scale features learned by our method and the great results of our deigned loss function prove that a promising approach for saliency detection is introduced in this paper. Experimental evaluations over six datasets demonstrate that our proposed method outperforms the previous state-of-the-art methods under different evaluation metrics. 

Based on the great performance and the real-time speed of our approach and its superiority over previous approaches, we plan to use our saliency detector in industrial object-related applications, such as object-based surveillance and object tracking. However, in real-world scenarios, images are affected by noise, which would lead to performance degradation of the most recently introduced saliency detectors~\cite{ijcai2018} including ours. Figure~\ref{fig:SM_Noise} shows the predicted saliency maps for images corrupted by Additive White Gaussian Noise (AWGN). As seen, our model fails to output accurate saliency predictions in the presence of noise. This motivates us to plan on enhancing the robustness of our method by handling noise in an end-to-end approach.



\bibliography{CAGNet_bib}

\end{document}